\pdfoutput=1
\documentclass[acmlarge, authorversion, screen, nonacm]{acmart}
\setcopyright{none}
\usepackage{amsmath,amsfonts}
\usepackage{graphicx}
\usepackage{algorithm,algorithmic}
\usepackage{hyperref}
\hypersetup{hidelinks=true}
\usepackage{textcomp}
\usepackage{rotating}
\usepackage{xcolor}
\usepackage{booktabs}
\usepackage{longtable}
\usepackage{dashrule}
\usepackage{array}
\usepackage{pifont}
\usepackage{multirow}
\usepackage{subcaption}
\usepackage{tikz}
\usetikzlibrary{arrows.meta,positioning,calc,fit,backgrounds}
\usepackage{fontawesome5}
\usepackage{xspace}
\usepackage{pgfplots}
\usepgfplotslibrary{colorbrewer}
\usepgfplotslibrary{statistics}
\usepackage[most,breakable]{tcolorbox}
\definecolor{ccqInk}{HTML}{2F3338}
\definecolor{ccqZone}{HTML}{F4F6F8}
\definecolor{ccqData}{HTML}{5A96C4}
\definecolor{ccqOpen}{HTML}{3E8C5C}
\tcbset{
  promptbox/.style={
    enhanced, breakable,
    colback=ccqZone, colframe=ccqZone, boxrule=0pt, arc=4pt,
    left=8pt, right=8pt, top=2pt, bottom=6pt,
    colbacktitle=ccqZone, coltitle=ccqInk, titlerule=0pt,
    toptitle=6pt, bottomtitle=0pt, fonttitle=\bfseries\small,
    fontupper=\small, coltext=ccqInk
  }
}
\newcommand{\ccqAgent}{\textcolor{ccqOpen}{\faIcon{robot}}\enspace}
\newcommand{\ccqSlot}[1]{\textcolor{ccqData}{\{#1\}}}

\begin{document}
\title{CCQ: A Multi-State Child Care Quality Dataset to Support AI for Children's Health Research}
\author{Victor Li}
\affiliation{%
  \institution{Emory University}
  \city{Atlanta}
  \state{GA}
  \country{USA}}
\email{victor.li@emory.edu}

\author{Yuzhang Xie}
\affiliation{%
  \institution{Emory University}
  \city{Atlanta}
  \state{GA}
  \country{USA}}
\email{yuzhang.xie@emory.edu}

\author{Ziwei Dong}
\affiliation{%
  \institution{Emory University}
  \city{Atlanta}
  \state{GA}
  \country{USA}}
\email{ziwei.dong@alumni.emory.edu}

\author{Qingyang Zhu}
\affiliation{%
  \institution{Emory University}
  \city{Atlanta}
  \state{GA}
  \country{USA}}
\email{qingyang.zhu@emory.edu}

\author{Wenjing Ma}
\affiliation{%
  \institution{Emory University}
  \city{Atlanta}
  \state{GA}
  \country{USA}}
\email{wenjing.ma@emory.edu}

\author{Carl Yang} 
\affiliation{%
  \institution{Emory University}
  \city{Atlanta}
  \state{GA}
  \country{USA}}
\email{j.carlyang@emory.edu}

\author{Jinbing Bai}
\affiliation{%
  \institution{Emory University}
  \city{Atlanta}
  \state{GA}
  \country{USA}}
\email{jinbing.bai@emory.edu}

\author{Huiwen Xu}
\affiliation{%
  \institution{Emory University}
  \city{Atlanta}
  \state{GA}
  \country{USA}}
\email{huiwen.xu@emory.edu}

\author{Jiaying Lu}\authornote{Corresponding author: Jiaying Lu, jiaying.lu@emory.edu}
\affiliation{%
  \institution{Emory University}
  \city{Atlanta}
  \state{GA}
  \country{USA}}
\email{jiaying.lu@emory.edu}

\begin{abstract}
High-quality child care in early life is a critical determinant of children’s growth and development. Research on child care quality has been constrained by fragmented, non-research-friendly, and privacy-bound datasets. We present \textit{CCQ} (Child Care Quality), a large-scale, de-identified dataset for applied data science research at the intersection of AI and early childhood health. \textit{CCQ} integrates 59,372 child care provider records across 12 U.S. states, covering diverse provider types as well as data schemas. To ensure research utility while protecting privacy, we implement an automated, LLM-based curation pipeline that anonymizes, cleans, and standardizes raw state records into two complementary releases: a cleaned textual release and a fully preprocessed tabular release. We also benchmark traditional machine learning models, tabular foundation models, and language models on quality rating prediction and important features analytics. Within a state, tabular classifiers on the preprocessed tables perform best. Across states, zero-shot transfer is near chance, but modest target-state supervision recovers most of the within-state performance, and pretraining on other states benefits finetuned language models. We release both datasets with all code to accelerate AI-driven research on child care quality and ultimately improve children's health and development.
\end{abstract}

\ccsdesc[500]{Applied computing~Health informatics}
\ccsdesc[300]{Computing methodologies~Machine learning}
\ccsdesc[300]{Information systems~Data cleaning}

\keywords{AI for children's health, child care quality rating prediction, benchmark and dataset, LLM-based data curation}

\renewcommand{\shortauthors}{V. Li, et al.}

\maketitle

\section{Introduction}
\label{sec:introduction}
Early-life environmental exposures, such as physical and social environments, affect children's health and development~\cite{blaisdell2022ECHO}. Among these, non-parental child care environments constitute a majority of sustained exposure. More than eight hours in care covers most of a young child's waking hours~\cite{cfoc2019}, and about 59\% of U.S. children age five and younger have at least one weekly non-parental care arrangement~\cite{cui2021ecpp}. Furthermore, approximately 15.7 million children in the U.S. under the age of five have routine child care arrangement~\cite{118hres1443ih}, where children spend an average of 36 hours a week in these settings~\cite{wood2013parents}. These early and prolonged experiences are linked to development well into adolescence: more hours of non-parental care before age four and a half predict more risk taking and impulsivity at age 15~\cite{vandell2010effects}. Higher-quality care is also associated with stronger cognitive and language skills by age three~\cite{national2000cog_lan_dev} and better cognitive and social development through second grade~\cite{peisnerfeinberg2001relation}. It also correlates with higher academic achievement and less externalizing behavior at age 15~\cite{vandell2010effects,vandell2023seccyd}. As a common part of family life, understanding the quality of child care is crucial for promoting children's health and development.

In order for families to act on these conclusions, accessible and relevant data on child care quality is needed. Provider-level quality ratings published through state-sponsored quality rating and improvement systems (QRIS) are designed to close this gap and inform parents' decisions~\cite{jenkins2021participates,markowitz2020simplifying}. Yet, parents do not use all the information available to them when judging quality~\cite{mocan2007lemons}. Specifically, in a study of low-income families, parents' evaluation of their child's program bore little correlation to these external measures~\cite{bassok2018parents}. However, since QRIS participation is voluntary in most states and coverage is often low, the providers a family is considering may simply have no rating~\cite{bassok2018parents,jenkins2021participates}. This coverage gap can be large: among the QRIS that reported participation in 2017, the median system rated only 60\% of licensed centers and 25\% of family child care programs, and New Jersey rated only 1\% of its centers~\cite{ecqa2019qris}. Beyond supporting parents, states use QRIS to help providers improve the quality of their care, and many pay higher subsidy rates to programs with higher ratings~\cite{jenkins2021participates,lin2024violations}. More complete coverage would make these efforts more effective as well. It would also advance research on how care quality relates to children's health and developmental outcomes, such as the NICHD Study of Early Child Care and Youth Development, which followed 1,364 children from birth and directly observed the child care settings they attended~\cite{vandell2023seccyd}, as well as research on the early-life environmental exposures examined by programs such as NIH ECHO~\cite{blaisdell2022ECHO}.

Machine learning offers a way to close the coverage gap by estimating quality for providers that have not been rated. In early care and education, it has been implemented to assess preschool teaching quality~\cite{li2022machine} and to predict the accreditation levels of early childhood institutions~\cite{noripansyah2023accreditation}, but each study used a small sample from a single rating system. A model fit to one rating system risks overfitting to that system's features and rating rules, and may not generalize to providers outside it, which is exactly the case where parents have no rating to rely on. Additionally, to support models that generalize across rating systems, the data must capture the many interconnected factors that shape quality. These include the physical environment (e.g., equipment, toys, facilities), food service (e.g., meals offered), policies and procedures (e.g., operational rules, parental access), program activities (e.g., curriculum, planned indoor/outdoor activities), and staffing (e.g., staff-to-child ratios)~\cite{cfoc2019}.

To our knowledge, no publicly available dataset features these factors with provider-level quality ratings across multiple rating systems. We attribute this gap to the challenges of collecting and preprocessing such heterogeneous data. Child care quality ratings (QR) are distributed by state across unstandardized QRIS, ranging from government-owned licensing databases and third-party websites to self-reported program descriptions~\cite{qcompendium,esplin2019self}. Each system stores and publishes its findings in its own schema~\cite{qcompendium}, which also contain missing values, errors, and outdated information~\cite{acf2024licensing}. In addition to these data limitations, preparing real-world children's health data often relies on costly expert annotation~\cite{kawatra2025minimal} and ad hoc, domain-specific preprocessing~\cite{lu2025PedCAFT} with teams of domain experts and data scientists~\cite{jacobson2024approaches}. Existing approaches for automated data curation pipelines also rely heavily on domain-specific knowledge\cite{narayan2022wrangle}, with hard-coded rules that make pipelines impractical to maintain, as well as, failing to generalize across diverse datasets~\cite{narayan2022wrangle}.

To fill this gap, we construct \textit{CCQ} (Child Care Quality), a curated dataset that combines provider-level quality ratings with provider features across twelve U.S. states. Because hand-written rules and expert annotation are time-consuming and labor-intensive at this scale, we leverage large language models (LLMs) to automatically collect, anonymize, and curate the raw state-specific records into two \textit{AI-ready} releases. They cover two families of models: a cleaned text release (\textit{raw}) that preserves the textual features of source records for language-based deep learning models, and a fully preprocessed tabular release (\textit{preprocessed}) that is strictly numerical and supports out-of-the-box analysis. Both releases are verified by hand for private information and data leakage, with the final values spot-checked against the source portals.

Using \textit{CCQ}, we benchmark models on two tasks of QR prediction, mirroring how state agencies would actually deploy them. In within-state prediction, a state already has many rated providers and wants to assign ratings to its remaining unrated providers under its own rating system. In cross-state prediction, a target state has few or no rated providers and must borrow from models trained on other states, whose schemas and rating systems differ. This harder setting tests the generalization across rating systems that single-system studies could not examine. We benchmark a representative set of models, spanning classical machine learning baselines, tabular deep learning and foundation models, and pretrained language models. First, we evaluate their predictive accuracy using ordinal agreement and balanced accuracy. Second, we apply feature attribution over the preprocessed features to identify which provider attributes drive predictions. Third, we measure cross-state transferability under different amount of labeled data, from zero-shot to 80\% of the target state's providers. Lastly, we examine whether full transfer learning in cross-state prediction improves prediction accuracy over the within-state setting. Together, these results establish reference baselines for the two QR prediction tasks and show where current models break down under realistic, heterogeneous conditions.

Our contributions are summarized as follows:
\begin{itemize}
    \item We construct \textit{CCQ}, a comprehensive provider-level dataset that combines child care quality ratings with provider features across twelve U.S. states. To our knowledge, it is the first publicly available, multi-state, AI-ready dataset of real-world child care quality ratings, supporting future AI research on children's health.
    \item We build an automated curation pipeline that leverages LLMs to efficiently collect, anonymize, and clean raw state-specific records into AI-ready releases. The pipeline reduces the time and labor that manual curation of multi-dimensional child care data typically requires. We verify every output by hand for private information and data leakage, and spot-check its values against the source portals.
    \item We benchmark 10 representative models, ranging from classical machine learning classifiers and tabular foundation models to finetuned language models, on within-state and cross-state QR prediction. We then use feature attribution to identify the provider attributes that drive predictions, and measure how cross-state transferability improves with more of target state labeled.
\end{itemize}

\begin{figure}[!htbp]
    \centering
    \resizebox{\linewidth}{!}{
\input{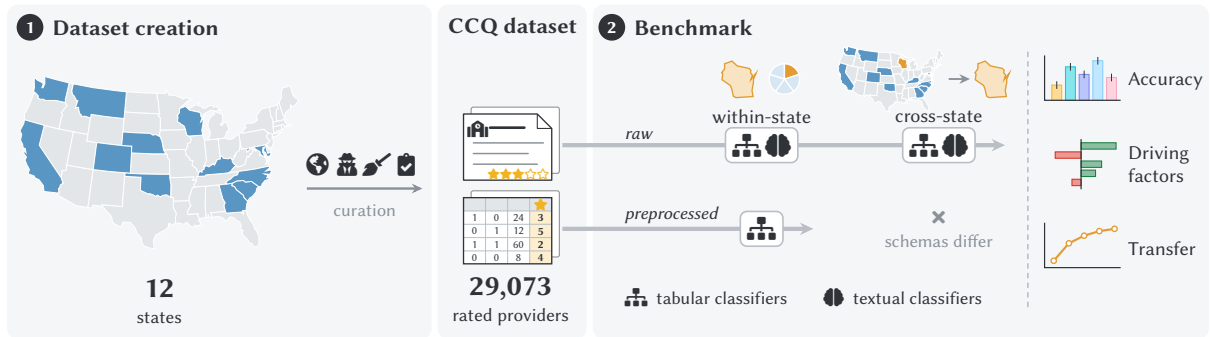}%
\begin{tikzpicture}[
  x=1cm, y=1cm,
  font=\sffamily\fontsize{7.4}{8.6}\selectfont,
  every node/.style={inner sep=0pt, outer sep=0pt, text=ccqInk},
  stage/.style={circle, fill=ccqInk, text=white, minimum size=10pt,
                font=\sffamily\bfseries\fontsize{6.8}{7}\selectfont},
  ttl/.style={anchor=west, font=\sffamily\bfseries\fontsize{8.4}{9}\selectfont},
  lbl/.style={align=center, font=\sffamily\fontsize{7.4}{8.4}\selectfont},
  note/.style={align=center, text=ccqMute, font=\sffamily\fontsize{6.8}{7.6}\selectfont},
  num/.style={align=center, font=\sffamily\fontsize{7}{7.8}\selectfont},
  ico/.style={font=\fontsize{8.6}{8.6}\selectfont},
  chip/.style={draw=ccqLine, fill=white, line width=0.6pt, rounded corners=2.5pt,
               inner xsep=2.6pt, inner ysep=2.2pt, font=\fontsize{8.6}{8.6}\selectfont},
  flow/.style={-{Stealth[length=4pt,width=3.6pt]}, draw=ccqMute, line width=0.9pt},
  row/.style={-{Stealth[length=5pt,width=4.6pt]}, draw=ccqLine, line width=1.6pt},
]
\node[stage] (n1) at (0.30,-0.30) {1};
\node[ttl] at ($(n1.east)+(0.12,0)$) {Dataset creation};

\ccqUSMap{shift={(2.00,-2.08)}, scale=0.39}{CA/ccqData, CO/ccqData, GA/ccqData, KY/ccqData,
  MD/ccqData, MT/ccqData, NC/ccqData, NE/ccqData, OK/ccqData, SC/ccqData, WA/ccqData, WI/ccqData}
\node[num] at (2.00,-3.90) {{\bfseries\fontsize{11}{11}\selectfont 12}\\[2pt]states};

\foreach \i/\ic in {0/globe-americas, 1/user-secret, 2/broom, 3/clipboard-check}{%
  \node[ico] at (4.08+\i*0.39,-2.10) {\faIcon{\ic}};}
\draw[flow] (3.94,-2.42) -- (5.46,-2.42);
\node[note] at (4.70,-2.70) {curation};

\node[ttl] at (5.80,-0.30) {CCQ dataset};
\pic at (6.62,-1.85) {ccq raw};
\pic at (6.62,-2.95) {ccq prep};
\node[num] at (6.62,-3.90) {{\bfseries\fontsize{11}{11}\selectfont 29,073}\\[2pt]rated providers};

\node[stage] (n2) at (7.95,-0.30) {2};
\node[ttl] at ($(n2.east)+(0.12,0)$) {Benchmark};

\ccqStateAt{WI}{(9.62,-0.98)}{0.48}{fill=ccqTgtLt, draw=ccqTgt, line width=0.5pt}
\pic at (10.22,-0.98) {ccq folds};
\node[lbl] at (9.92,-1.47) {within-state};

\ccqUSMap{shift={(11.62,-0.98)}, scale=0.148}{CA/ccqData, CO/ccqData, GA/ccqData, KY/ccqData,
  MD/ccqData, MT/ccqData, NC/ccqData, NE/ccqData, OK/ccqData, SC/ccqData, WA/ccqData, WI/ccqTgt}
\draw[flow, line width=0.7pt] (12.38,-0.98) -- (12.66,-0.98);
\ccqStateAt{WI}{(12.95,-0.98)}{0.48}{fill=ccqTgtLt, draw=ccqTgt, line width=0.5pt}
\node[lbl] at (12.25,-1.47) {cross-state};

\draw[row] (7.30,-1.85) -- (13.12,-1.85);
\draw[row] (7.30,-2.95) -- (10.60,-2.95);
\node[note, anchor=south west, text=ccqInk] at (8.12,-1.77) {\itshape raw};
\node[note, anchor=south west, text=ccqInk] at (8.12,-2.87) {\itshape preprocessed};
\node[chip] at (9.92,-1.85) {\faIcon{sitemap}\,\faIcon{brain}};
\node[chip] at (12.25,-1.85) {\faIcon{sitemap}\,\faIcon{brain}};
\node[chip] at (9.92,-2.95) {\faIcon{sitemap}};
\node[note] at (12.25,-2.95) {\faIcon{times}\\[1.2pt]schemas differ};

\node[note, text=ccqInk, anchor=west] at (8.12,-3.85)
  {\faIcon{sitemap}\ \ tabular classifiers\qquad\faIcon{brain}\ \ textual classifiers};

\draw[ccqLine, line width=0.6pt, dash pattern=on 2pt off 1.6pt] (13.42,-0.62) -- (13.42,-4.02);
\pic at (14.10,-1.00) {ccq bars};
\node[lbl, anchor=west] at (14.74,-1.00) {Accuracy};
\pic at (14.10,-2.10) {ccq shap};
\node[lbl, anchor=west, align=left] at (14.74,-2.10) {Driving\\factors};
\pic at (14.10,-3.20) {ccq curve};
\node[lbl, anchor=west] at (14.74,-3.20) {Transfer};

\begin{scope}[on background layer]
  \fill[ccqZone, rounded corners=4pt] (0,0) rectangle (5.58,-4.40);
  \fill[ccqZone, rounded corners=4pt] (5.66,0) rectangle (7.62,-4.40);
  \fill[ccqZone, rounded corners=4pt] (7.70,0) rectangle (15.80,-4.40);
\end{scope}
\end{tikzpicture}%
}
    \caption{Overview of the \textit{CCQ} study. \textbf{Stage 1} curates the QRIS records of twelve states in four steps: collection, anonymization, cleaning, and verification (Figure~\ref{fig:dataset_pipeline}). The \textit{raw} release retains textual feature while the \textit{preprocessed} release is strictly numerical. \textbf{Stage 2} benchmarks tabular classifiers and language models on two tasks. Within-state prediction uses five-fold cross-validation and cross-state prediction holds out one state (orange) and trains on the others (blue). Because the \textit{preprocessed} data schemas differ across states, it is only used for within-state prediction. The trained models are analyzed for prediction accuracy, feature attribution, and cross-state transferability.}
    \label{fig:study_overview}
\end{figure}
\section{Related Work}
\label{sec:related_work}

\subsection{Child Care Quality Datasets}
Though several child care quality datasets~\cite{acecqa,albertadata,banpdm,faces2014} are publicly available and could support quality prediction modeling, they are limited in their coverage and in the features they record compared to \textit{CCQ}. The Australian Children's Education \& Care Quality Authority publishes the ratings of more than 17,000 early childhood services, all assessed under the National Quality Standard~\cite{acecqa,acecqa2026snapshot}. Indonesia's national accreditation agency publishes the accreditation scores for more than 174,000 early childhood education institutions~\cite{banpdm,pusdatin2026paud}. Beyond the ratings themselves, however, both record little about the factors that shape quality. The Australian dataset records each service's provider, location, care types, number of approved places, and operating hours~\cite{acecqa}. The Indonesian dataset generally records only each institution's accreditation score~\cite{banpdm}. On the other hand, the Canadian province of Alberta releases quarterly records of 2,720 licensed child care providers with their location, type, capacity, and inspection results, but assigns them no quality ratings~\cite{albertadata}. The U.S. Head Start Family and Child Experiences Survey~\cite{faces2014} also records observed quality scores (CLASS and ECERS-R) for 669 classrooms in 347 Head Start centers~\cite{aikens2016tracking}, but assigns no quality rating. Lastly, the Quality Compendium~\cite{qcompendium} catalogs QRIS programs across U.S. states, but it is curated at the system level rather than the provider level. To our knowledge, no unified, multi-state, provider-level QRIS dataset is publicly available. \textit{CCQ} fills this gap by assembling provider-level records and quality ratings across twelve U.S. states into a single, consistently curated dataset with features that span the factors shaping care quality~\cite{cfoc2019}, although each state's schema differs.

\subsection{Child Care Quality Prediction}
Predictive modeling of child care and early childhood education quality has not kept up with modern machine learning. Existing studies rely mostly on regression models and classical classifiers~\cite{goelman2006towards,doherty2006predictors,aikens2016tracking,siraj2019comparing,li2022machine,char2023macro,noripansyah2023accreditation,harrison2024structures}. To our knowledge, no study evaluates tree ensembles, tabular foundation models, or language models. Char et al.~\cite{char2023macro} and Harrison et al.~\cite{harrison2024structures} analyzed the Australian national register of quality ratings~\cite{acecqa} using multinomial logistic regression. Siraj et al.~\cite{siraj2019comparing} found correlations between these ratings and observed classroom quality using hierarchical multiple regression. Aikens et al.~\cite{aikens2016tracking} analyzed Head Start classroom quality in the FACES dataset~\cite{faces2014} using linear and logistic regression. Noripansyah et al.~\cite{noripansyah2023accreditation} compared support vector classifiers (SVC), K-Nearest Neighbors (KNN), and neural network approaches for predicting the accreditation scores of 225 early childhood education providers in South Kalimantan, Indonesia~\cite{banpdm,pusdatin2026paud}. Other studies rely on samples that the researchers collected themselves. Goelman et al.~\cite{goelman2006towards} developed a path-analytic prediction model of center-based child care quality across 239 Canadian centers, while Doherty et al.~\cite{doherty2006predictors} used hierarchical regression to predict the quality of 231 Canadian family child care providers. These studies all share two limitations that our benchmark addresses. First, each study fits and evaluates models using data from a single program with a limited number of labeled providers, so the reported performance does not reflect how models behave on large, heterogeneous datasets. Second, no study examines whether a model trained on one rating system transfers to another. In contrast, we evaluate a more comprehensive set of conventional machine learning methods, tabular foundation models, and deep learning models on much larger datasets. We also explicitly study the heterogeneous cross-state transfer setting, in which models are trained on several states' data despite differing schemas, which reflects the fragmentation of real-world child care rating programs.
\section{Dataset Creation}
\label{sec:dataset_creation}

\subsection{Data Sources}
\label{sec:data-sources}
The \textit{CCQ} (Child Care Quality) dataset integrates publicly available child care data across 12 U.S. states: California (CA), Colorado (CO), Georgia (GA), Kentucky (KY), Maryland (MD), Montana (MT), North Carolina (NC), Nebraska (NE), Oklahoma (OK), South Carolina (SC), Washington (WA), and Wisconsin (WI). In total, we assemble 59,372 records, originally released by state agencies or their designated partners. Each state agency operates its own QRIS, which differ in design but share the purpose of helping families make informed decisions~\cite{jenkins2021participates,markowitz2020simplifying}. Table~\ref{tab:data_source} lists the source agency, QRIS, QR scale, and provider counts for each state. With the exception of Georgia, which uses a 1--3 scale, all states publish a 1--5 QR scale. Overall, only 29,073 providers (49.0\%) in the dataset possess an assigned QR score, highlighting the need for predictive models that can estimate ratings for the remaining facilities.

\begin{table}[!htbp]
\centering
\caption{Data sources overview. \#Feature includes the provider ID and QR score.}
\label{tab:data_source}
\resizebox{\linewidth}{!}{
\begin{tabular}{lcccccc}
\toprule
\textbf{State} & \textbf{Data Source} &\textbf{\#Provider} & \textbf{\#Rated} & \textbf{QRIS} & \textbf{QR Scale} & \textbf{\#Feature}\\
\midrule
CA & CA Child Care Resource \& Referral Network~\cite{CACCQ}& 10,715 & 847 & Quality Counts California  & 1--5 & 142\\
CO & CO Dept of Early Childhood~\cite{COCCQ} & 4,508  & 3,423 & Colorado Shines Rating & 1--5 & 64\\
GA & GA Dept of Early Care and Learning~\cite{GACCQ} & 8,192  & 2,897 & Quality Rated & 1--3 & 203\\
KY & KY Cabinet for Health and Family Services~\cite{KYCCQ} & 1,929  & 1,896 & Kentucky All STARS & 1--5 & 61\\
MD & MD State Dept of Education~\cite{MDCCQ} & 7,106  & 5,005 & Maryland EXCELS  & 1--5 & 32\\
MT & MT Dept of Public Health \& Human Services~\cite{MTCCQ} & 213    & 181 & Best Beginnings STARS to Quality  & 1--5 & 12\\
NC & NC Div. of Child Development \& Early Education~\cite{NCCCQ}  & 3,264  & 3,050 & NC Star Rated License& 1--5 & 30\\
NE &  NE Dept of Education~\cite{NECQ} & 3,274  & 1,079 & Step Up to Quality & 1--5 & 38\\
OK &  OK Dept of Human Services~\cite{OKCCQ} &  2,557  & 2,554 & Reaching for the Stars & 1--5 & 179\\
SC &  SC Dept of Social Services~\cite{SCCCQ} & 2,400 & 1,223 & ABC Quality & 1--5 & 10\\
WA & WA Dept of Children, Youth, \& Families~\cite{WACCQ} & 10,399 & 3,168 &   Early Achievers & 1--5 & 169\\
WI &  WI Dept of Children and Families~\cite{WICCQ} & 4,815  & 3,750 & YoungStar & 1--5 & 164\\
\midrule
Total & -- & 59,372 & 29,073 & -- & -- & --\\
\bottomrule
\end{tabular}
}

\end{table}

\subsection{Construction Pipeline}
\label{sec:construction-pipeline}
Since each state administers and publishes its own QRIS~\cite{qcompendium}, there is no federal registry, common schema, or common format for the data. The practical consequence is that data from each state require substantial state-specific collection and preprocessing. We believe this lack of standardization is a key reason why no national-scale dataset of provider-level quality ratings currently exists. Our construction pipeline uses Georgia as a human-implemented reference that the remaining states are adapted from. It contains four main steps: collection, anonymization, cleaning, and verification. We chose Georgia as the reference because it has the most published fields of the 12 states (Table~\ref{tab:data_source}), which all publish its data on easy accessible portals.

\begin{figure}[!htbp]
    \centering
    \resizebox{\linewidth}{!}{\input{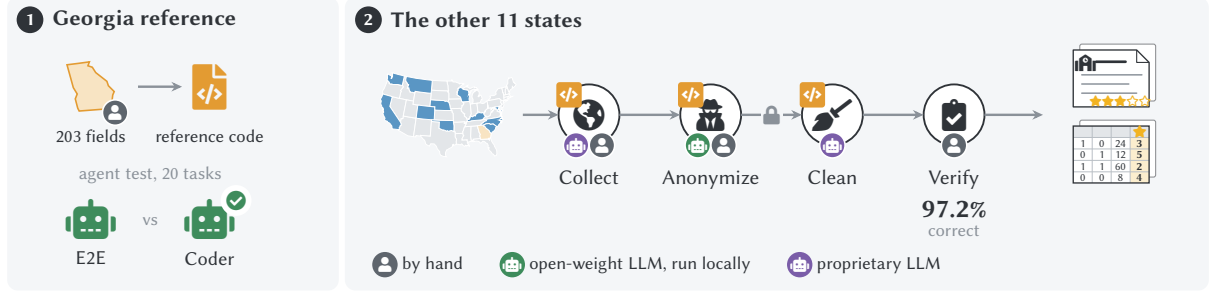}%
\begin{tikzpicture}[
  x=1cm, y=1cm,
  font=\sffamily\fontsize{7.4}{8.6}\selectfont,
  every node/.style={inner sep=0pt, outer sep=0pt, text=ccqInk},
  stage/.style={circle, fill=ccqInk, text=white, minimum size=10pt,
                font=\sffamily\bfseries\fontsize{6.8}{7}\selectfont},
  ttl/.style={anchor=west, font=\sffamily\bfseries\fontsize{8.4}{9}\selectfont},
  lbl/.style={align=center, font=\sffamily\fontsize{7.4}{8.4}\selectfont},
  stepname/.style={align=center, font=\sffamily\fontsize{7.8}{8.8}\selectfont},
  note/.style={align=center, text=ccqMute, font=\sffamily\fontsize{6.8}{7.6}\selectfont},
  ico/.style={font=\fontsize{8.6}{8.6}\selectfont},
  gacode/.style={rounded corners=1.2pt, fill=ccqTgt, text=white, minimum width=9.5pt,
               minimum height=9.5pt, inner sep=0pt, font=\fontsize{5.4}{5.4}\selectfont},
  flow/.style={-{Stealth[length=4pt,width=3.6pt]}, draw=ccqMute, line width=0.9pt},
]
\node[stage] (n1) at (0.30,-0.30) {1};
\node[ttl] at ($(n1.east)+(0.12,0)$) {Georgia reference};

\ccqStateAt{GA}{(1.10,-1.18)}{0.64}{fill=ccqTgtLt, draw=ccqTgt, line width=0.6pt}
\node[who=ccqHand, draw=white, line width=0.6pt] at (1.42,-1.52) {\faIcon{user}};
\draw[flow] (1.72,-1.18) -- (2.30,-1.18);
\node[text=ccqTgt, font=\fontsize{17}{17}\selectfont] at (2.66,-1.18) {\faIcon{file-code}};
\node[note, text=ccqInk] at (1.10,-1.82) {203 fields};
\node[note, text=ccqInk] at (2.66,-1.82) {reference code};

\node[note] at (1.88,-2.34) {agent test, 20 tasks};
\node[text=ccqOpen, font=\fontsize{15}{15}\selectfont] (e2e) at (1.10,-2.92) {\faIcon{robot}};
\node[text=ccqOpen, font=\fontsize{15}{15}\selectfont] (cod) at (2.66,-2.92) {\faIcon{robot}};
\node[note] at (1.88,-2.94) {vs};
\node[lbl] at (1.10,-3.43) {E2E};
\node[lbl] at (2.66,-3.43) {Coder};
\node[text=ccqOpen, font=\fontsize{7.5}{7.5}\selectfont, fill=white, circle, inner sep=0.4pt]
  at ($(cod.north east)+(0.02,-0.02)$) {\faIcon{check-circle}};

\node[stage] (n2) at (4.75,-0.30) {2};
\node[ttl] at ($(n2.east)+(0.12,0)$) {The other 11 states};

\ccqUSMap{shift={(5.80,-1.55)}, scale=0.19}{CA/ccqData, CO/ccqData, GA/ccqTgtLt, KY/ccqData,
  MD/ccqData, MT/ccqData, NC/ccqData, NE/ccqData, OK/ccqData, SC/ccqData, WA/ccqData, WI/ccqData}

\foreach \x/\ic/\name in {7.65/globe-americas/Collect, 9.25/user-secret/Anonymize,
                          10.85/broom/Clean, 12.45/clipboard-check/Verify}{%
  \node[station] (st\name) at (\x,-1.55) {\faIcon{\ic}};
  \node[stepname, anchor=base] at (\x,-2.46) {\name};}
\draw[flow] (6.78,-1.55) -- (stCollect.west);
\draw[flow] (stCollect.east) -- (stAnonymize.west);
\draw[flow] (stAnonymize.east) -- (stClean.west);
\draw[flow] (stClean.east) -- (stVerify.west);

\node[who=ccqProp, draw=white, line width=0.6pt] at ($(stCollect.south)+(-0.17,0)$)   {\faIcon{robot}};
\node[who=ccqHand, draw=white, line width=0.6pt] at ($(stCollect.south)+(0.17,0)$)    {\faIcon{user}};
\node[who=ccqOpen, draw=white, line width=0.6pt] at ($(stAnonymize.south)+(-0.17,0)$) {\faIcon{robot}};
\node[who=ccqHand, draw=white, line width=0.6pt] at ($(stAnonymize.south)+(0.17,0)$)  {\faIcon{user}};
\node[who=ccqProp, draw=white, line width=0.6pt] at (stClean.south)                  {\faIcon{robot}};
\node[who=ccqHand, draw=white, line width=0.6pt] at (stVerify.south)                 {\faIcon{user}};
\foreach \s in {Collect, Anonymize, Clean}{%
  \node[gacode] at ($(st\s.north west)+(0.02,-0.02)$) {\faIcon{code}};}
\node[fill=ccqZone, inner sep=1pt, text=ccqMute, font=\fontsize{7}{7}\selectfont]
  at ($(stAnonymize.east)!0.5!(stClean.west)$) {\faIcon{lock}};

\node[font=\sffamily\bfseries\fontsize{10}{10}\selectfont] at (12.45,-2.78) {97.2\%};
\node[note] at (12.45,-3.08) {correct};

\draw[flow] (stVerify.east) -- (13.62,-1.55);
\pic[scale=0.86] at (14.55,-1.05) {ccq raw};
\pic[scale=0.86] at (14.55,-2.05) {ccq prep};

\node[anchor=west, note, text=ccqInk] at (4.80,-3.52)
  {\tikz[baseline=-0.6ex]\node[who=ccqHand]{\faIcon{user}};\ by hand\qquad
   \tikz[baseline=-0.6ex]\node[who=ccqOpen]{\faIcon{robot}};\ open-weight LLM, run locally\qquad
   \tikz[baseline=-0.6ex]\node[who=ccqProp]{\faIcon{robot}};\ proprietary LLM};

\begin{scope}[on background layer]
  \fill[ccqZone, rounded corners=4pt] (0,0) rectangle (4.36,-3.86);
  \fill[ccqZone, rounded corners=4pt] (4.44,0) rectangle (15.80,-3.86);
\end{scope}
\end{tikzpicture}%
}
    \caption{The \textit{CCQ} construction pipeline. \textbf{(1)} The Georgia reference is implemented by hand. Its human-curated data are the ground truth for testing two agents on twenty curation tasks: a per-sample agent (E2E) and a code-writing agent (Coder), which is adopted. \textbf{(2)} The other eleven states are collected, anonymized, and cleaned with the help of LLMs. Human verification at each step is shown with icons indicating who performed each step. The orange code icon indicate when Georgia's code is used as and in-context example. Lastly, two annotators spot-checked the curated data against the live portals, finding 97.2\% of the checked values correct.}
    \label{fig:dataset_pipeline}
\end{figure}

\subsubsection{Collection}
\label{sssec:data_curation-collect}
For every state, we collect a single table where rows represent providers and columns represent provider attributes. The first two columns are always the provider identifier (ID) and the QR score. In the hand-written collection pipeline for Georgia, providers are enumerated through the state agency's search portal, fetching all available information. We then prompted a coding agent (Claude~\cite{anthropic2025claude4}) to adapt that reference to each remaining state, following prior work that uses LLMs to generate web scrapers~\cite{autoscraper} and LM agents to write code~\cite{yang2024sweagent}. Each state's collected columns were then verified by hand, using the column names and their unique values, confirming that every informative attribute available had been recorded. 

\subsubsection{Cleaning}
\label{sssec:data_curation-clean}
To support different machine learning models, every state is cleaned to produce two separate releases of the data: \textit{raw} and \textit{preprocessed}. The \textit{raw} release applies minimal preprocessing and preserves the tabular text of the source records, so that each provider record can be directly fed as natural language into text-based models. On the other hand, the \textit{preprocessed} release applies a complete preprocessing pipeline for out-of-the-box analysis, creating a strictly numerical dataset. Both releases remove sensitive private columns as well as data-leakage columns. The private columns are those that identify a provider or a person rather than describe the program, such as provider names, staff names, addresses, emails, phone numbers, and websites. Removing them is a de-identification step in the sense used by NIST, which covers establishments as well as individuals~\cite{garfinkel2023nist800188}. This also includes free-text narratives some states publish, such as inspection, complaint, and violation reports, which carry private information about the provider. Coarse geography, county, and region are retained. The data-leakage columns are those that disclose the rating rather than predict it, such as award dates or different formats of rating scores. Missing values are left empty in the \textit{raw} release, but encoded explicitly in the \textit{preprocessed} release. Lastly, unrated providers are also removed.

As with collection, Georgia serves as the reference, and its preprocessing was carried out under strict expert supervision. Against that hand-curated ground truth we then measured two zero-shot agents on twenty curation tasks for anonymization and cleaning: a per-sample agent (\textit{Qwen3-8B}~\cite{yang2025qwen3}, ``E2E''), prompted once per record, and a code-writing agent (\textit{Qwen3-Coder-30B}~\cite{qwen3_coder_30b_a3b_instruct}, ``Coder''), prompted to write one reusable cleaning function per field with the ground-truth code withheld. Table~\ref{tab:llm_summary_eval} reports the results, grouping the twenty tasks into five types. Each row averages over the tasks of one type, and the \#Tasks column gives their number.

Although the coding agent does not achieve perfect performance, we adopt it to conduct anonymization and generate cleaning scripts. Its accuracy is comparable to the per-sample agent, except on rare edge cases within the data, at a fraction of the cost. To remedy these edge cases, we include the Georgia pipeline as an in-context example in every anonymization and cleaning prompt. The in-context example is always code: the reference implementation's anonymization and cleaning functions are supplied to the agent, and no provider record from any state is ever placed in the prompt. We also use a proprietary LLM (Claude~\cite{anthropic2025claude4}), whose code-generation ability lets us produce an entire state's pipeline in a single pass rather than one cleaning function at a time. Importantly, no private information was exposed to the proprietary model as anonymization was performed locally with open-source models beforehand, and every column flagged as sensitive was removed by hand before any cleaning script was generated.

\begin{table}[!htbp]
\centering
\footnotesize
\caption{LLM-based curation results on the reference state Georgia. Token counts are per provider record, so the code-writing agent's one-time cost is amortized over the 8,192 records in the Georgia reference implementation.}
\label{tab:llm_summary_eval}
\begin{tabular}{llcccc}
\toprule
\textbf{Curation Task} & \textbf{Model} & \textbf{\#Tasks} & \textbf{Acc\%} & \textbf{In Tok.} & \textbf{Out Tok.}\\
\midrule
\multicolumn{6}{l}{\textit{Anonymization}} \\
Column removal & E2E & 1 & $97.12$ & $71.0$ & $1.0$ \\
Column removal & Coder & 1 & $99.28$ & $103.1$ & $1.0$ \\
De-identification & E2E & 3 & $35.2 \pm 42.8$ & $107.2\pm 22.6$ & $4.4\pm 2.5$ \\
De-identification & Coder & 3 & $100.0 \pm0.0$ & $0.03\pm 0.00$ & $0.05\pm 0.01$\\
\midrule
\multicolumn{6}{l}{\textit{Cleaning}} \\
Text $\to$ numeric & E2E & 5 & $93.3\pm 13.4$ & $165.3\pm 18.4$ & $2.7\pm 1.2$\\
Text $\to$ numeric & Coder & 5 & $99.98\pm 0.02$ & $0.14\pm0.12$ & $0.03\pm0.01$ \\
Text $\to$ multi-label & E2E & 8 & $97.2\pm3.8$ & $390.1\pm103.3$ & $22.2\pm 9.0$\\
Text $\to$ multi-label & Coder & 8 & $88.9\pm22.1$ & $0.53\pm0.96$ & $0.08\pm0.03$ \\
Compound field & E2E & 3 & $95.8\pm 3.0$ & $511.3\pm 345.4$ & $13.9\pm12.0$ \\
Compound field & Coder & 3 & $90.2\pm11.9$ & $1.92\pm1.29$ & $0.08\pm 0.03$\\
\bottomrule
\end{tabular}

\end{table}

The following boxes give one prompt from each agent. The anonymization prompt asks the per-sample agent for a single privacy decision about one column, given the column name and a sample of that column's values. The cleaning prompt asks the code-writing agent for one reusable function, given the field name and the field's unique values, with the requirements stated as numbered constraints. The remaining prompts follow these two shapes and vary only in the operation requested and the output columns written.

\begin{tcolorbox}[promptbox, title={\ccqAgent Prompt of the E2E Agent for Anonymization: Column Removal}]
You are an AI researcher studying child care quality data. In order to support open science, you need to determine which fields in the original data need to be anonymized to protect privacy. Given the variable name ``\ccqSlot{PLACEHOLDER\_NAME}'' and sample associated values ``\ccqSlot{PLACEHOLDER\_VAL1, PLACEHOLDER\_VAL2, ...}'', respond ``Yes'' if it contains protected privacy information, ``No'' if not.
\end{tcolorbox}

\begin{tcolorbox}[promptbox, title={\ccqAgent Prompt of the Coder Agent for Cleaning: Compound Field Split}]
You are an AI research assistant creating a child care quality dataset. In order to allow future researchers to easily use this data, some of the fields in the raw data require cleaning. Given the field name `\ccqSlot{PLACEHOLDER\_FIELD\_NAME}' and unique values `\ccqSlot{PLACEHOLDER\_UNIQUE\_VALUES}', write a min-max split function with the following requirements.\\
1. The function should have two inputs: `csv\_file\_path' and `column\_name'.\\
2. The function should first read the `csv\_file\_path' into a pandas dataframe and then extract the `column\_name' column from it.\\
3. The function should then clean the column by splitting the input into two separate values: min and max. Then, remove any leading non-numerical prefix (symbols/words/spaces) from each value. If the input is nan, missing, unknown (i.e., it is np.nan, `nan', `NaN', `None', `null', `unknown', an empty string, or whitespace-only), set the cleaned value to be None.\\
4. After splitting the input into min and max, replace the original column with two new columns titled `predicted\_min' and `predicted\_max' and save the new dataframe to the original `csv\_file\_path'.\\
Now, output the ONLY the code for the function. Do not output any extra text or explanations.
\end{tcolorbox}
\vspace{0.5em}

\subsubsection{Verification}
\label{sec:data-verification}
Human verification effort was allocated for efficiency under time constraints. First, the Georgia reference implementation code was fully verified for collection, anonymization, and cleaning. For the other states, only the anonymization step was fully verified by hand. Both critical failure modes, privacy and data leakage, were checked to ensure that no text or derived feature contains any private or quality-rating information. However, the generated cleaning scripts were not verified line by line for correctness. Instead, we spot check the correctness of the end-to-end output, from collection to cleaned data, through a human review against the live source portals.

To estimate end-to-end collection and cleaning error, two authors independently re-derived a small set of fields by hand from each sampled provider's page on the state portal and compared them to the released values. We reviewed 10 of the 12 states. In each, we drew 20 rated providers at random, stratified by quality rating in proportion to the state's rating distribution. Georgia is excluded because it is the reference implementation and Montana because its quality ratings originate from a standardized database. For each of the 200 providers (0.69\% of the 29,073 rated providers), we checked the four most common and critical fields: quality rating, provider type, licensed capacity, and ages served. Since not all of these fields exist in every state, we exclude each state-field pair where the field is absent. This leaves 740 checked values per annotator.

Table~\ref{tab:data_verfication} reports the results per state, and its last row (ALL) pools them over the ten states. The two annotators judged 96.8\% and 97.6\% of the released values correct, or 1,438 of the 1,480 checks combined (97.2\%). They agreed with each other on 730 of the 740 values. Pooled over states, Cohen's kappa~\cite{cohen1968qwk} is 98.1 for quality rating, 99.4 for provider type, 95.9 for licensed capacity, and 99.4 for ages served, all within the range of almost perfect agreement~\cite{landis1977measurement}. Agreement is lowest for licensed capacity in Kentucky (68.0). Five of the ten disagreements are Kentucky capacities that one annotator flagged and the other accepted. Furthermore, there are 16 values that both annotators flagged (2.2\%). Four are provider types in Colorado, where both annotators found 4 of the 20 values wrong. Five are quality ratings, one each in Kentucky, Nebraska, and Wisconsin and two in Washington. The remaining seven are four licensed capacities and three ages-served values. We attribute these mismatches to recently updated records that changed between time of collection and verification. However, we did not explicitly determine the cause of each mismatch.

\begin{table}[!htbp]
    \centering
    \small
    \setlength{\tabcolsep}{3.5pt}
    \caption{Manual data verification results per state (20 providers per state). H1 and H2 give the percentage of released values each annotator judged correct against the live portal. $\kappa$ is Cohen's kappa ($\times 100$) between their correct/incorrect judgments. The ALL row pools the states that publish each field. ``NA'' indicates that the state does not publish the field.}
    \label{tab:data_verfication}
    \begin{tabular}{l|ccc|ccc|ccc|ccc}
\toprule
\multirow{2}{*}{\textbf{State}} & \multicolumn{3}{c|}{\textbf{QR score}} & \multicolumn{3}{c|}{\textbf{Provider type}} & \multicolumn{3}{c|}{\textbf{Licensed capacity}} & \multicolumn{3}{c}{\textbf{Ages served}} \\
 & H1 & H2 & $\kappa$ & H1 & H2 & $\kappa$ & H1 & H2 & $\kappa$ & H1 & H2 & $\kappa$ \\
\midrule
CA & 100.0 & 100.0 & 100.0 & NA & NA & NA & 100.0 & 100.0 & 100.0 & 95.0 & 95.0 & 100.0 \\
CO & 100.0 & 100.0 & 100.0 & 80.0 & 80.0 & 91.3 & 90.0 & 95.0 & 94.6 & 100.0 & 100.0 & 100.0 \\
KY & 90.0 & 95.0 & 91.6 & 100.0 & 100.0 & 100.0 & 70.0 & 95.0 & 68.0 & NA & NA & NA \\
MD & 100.0 & 100.0 & 100.0 & 100.0 & 100.0 & 100.0 & 95.0 & 95.0 & 100.0 & 100.0 & 100.0 & 100.0 \\
NC & 100.0 & 95.0 & 92.4 & 100.0 & 100.0 & 100.0 & 100.0 & 100.0 & 100.0 & 100.0 & 100.0 & 100.0 \\
NE & 95.0 & 95.0 & 100.0 & 100.0 & 100.0 & 100.0 & 100.0 & 100.0 & 100.0 & 100.0 & 100.0 & 100.0 \\
OK & 100.0 & 100.0 & 100.0 & 100.0 & 100.0 & 100.0 & 100.0 & 100.0 & 100.0 & 100.0 & 100.0 & 100.0 \\
SC & 100.0 & 100.0 & 100.0 & 100.0 & 100.0 & 100.0 & 100.0 & 100.0 & 100.0 & NA & NA & NA \\
WA & 90.0 & 85.0 & 89.1 & 100.0 & 100.0 & 100.0 & 95.0 & 100.0 & 93.8 & 100.0 & 100.0 & 100.0 \\
WI & 95.0 & 95.0 & 100.0 & 100.0 & 100.0 & 100.0 & 95.0 & 95.0 & 100.0 & 90.0 & 90.0 & 94.3 \\
\midrule
ALL & 97.0 & 96.5 & 98.1 & 97.8 & 97.8 & 99.36 & 94.5 & 98.0 & 95.87 & 98.1 & 98.1 & 99.35 \\
\bottomrule
\end{tabular}

\end{table}

\subsection{Dataset Overview}
\label{sec:dataset-overview}
\textit{CCQ} contains 59,372 child care providers drawn from 12 states: four in the West (CA, CO, MT, WA), two in the Midwest (NE, WI), and six in the South (GA, KY, MD, NC, OK, SC). Records span learning centers, family child care homes, and other specialized provider types. Of the 59,372 providers, 29,073 (49.0\%) carry a published quality rating and form the labeled portion of the dataset. The size of each state's dataset differs greatly (Table~\ref{tab:data_source}), from 181 rated providers in Montana to 5,005 in Maryland. Additionally, feature schemas are defined state by state and unique, ranging from 10 attributes in South Carolina to 203 in Georgia, with a median of 62.5 attributes.

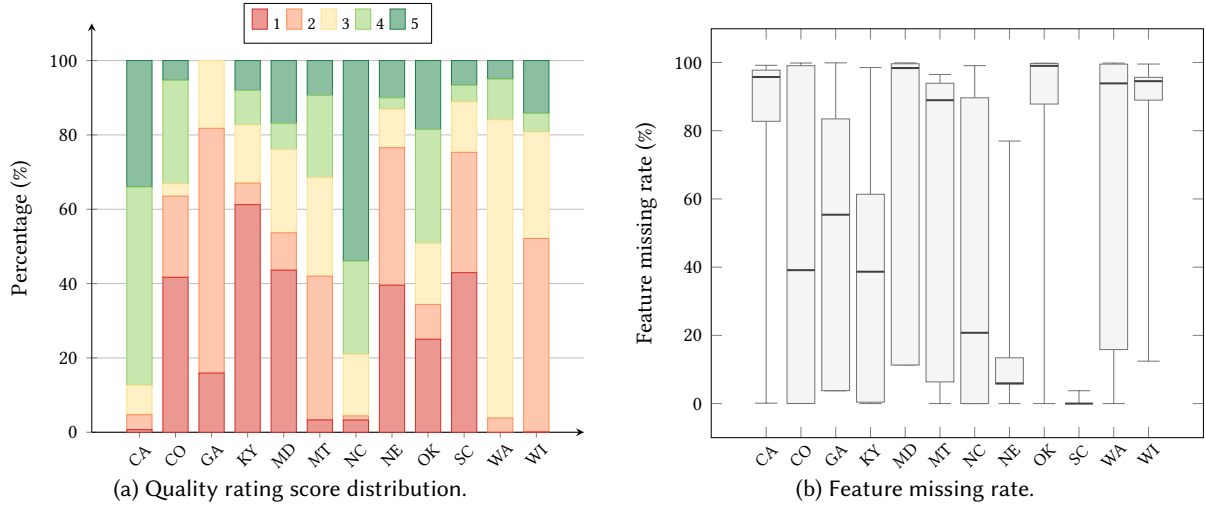
\begin{figure}[!htbp]
    \centering
    \begin{subfigure}[b]{0.48\textwidth}
    \centering
    \definecolor{qr1}{HTML}{D73027}
\definecolor{qr2}{HTML}{FC8D59}
\definecolor{qr3}{HTML}{FEE08B}
\definecolor{qr4}{HTML}{91CF60}
\definecolor{qr5}{HTML}{1A7F45}
\begin{tikzpicture}[scale=0.95]
  \begin{axis}[
    axis lines=left,
    ymajorgrids,
    ybar stacked,
    ymin=0,
    ymax=100,
    enlarge y limits={upper=0.12},
    enlarge x limits=0.12,      
    legend style={at={(0.5,1.05)},
      anchor=north,legend columns=-1},
    ylabel={Percentage (\%)},
    ylabel near ticks,
    ylabel style={font=\small},
    y tick label style={font=\footnotesize},
    symbolic x coords={CA, CO, GA, KY, MD, MT, NC, NE, OK, SC, WA, WI},
    xtick=data,
    x tick label style={yshift=-7pt, rotate=45, anchor=center, font=\scriptsize},
    legend style={font=\scriptsize},
    ]
\addplot[ybar, fill=qr1!50, draw=qr1] plot coordinates {(CA,0.71) (CO,41.69) (GA,15.91) (KY,61.23) (MD,43.60) (MT,3.31) (NC,3.28) (NE,39.57) (OK,25.02) (SC,42.93) (WA,0.00) (WI,0.05)};
\addplot[ybar, fill=qr2!50, draw=qr2] plot coordinates {(CA,4.01) (CO,21.82) (GA,65.86) (KY,5.80) (MD,10.05) (MT,38.67) (NC,1.11) (NE,36.98) (OK,9.36) (SC,32.38) (WA,3.82) (WI,52.08)};
\addplot[ybar, fill=qr3!50, draw=qr3] plot coordinates {(CA,7.91) (CO,3.36) (GA,18.23) (KY,15.66) (MD,22.40) (MT,26.52) (NC,16.59) (NE,10.38) (OK,16.41) (SC,13.57) (WA,80.24) (WI,28.67)};
\addplot[ybar, fill=qr4!50, draw=qr4] plot coordinates {(CA,53.36) (CO,27.84) (GA,0.00) (KY,9.28) (MD,6.99) (MT,22.10) (NC,25.08) (NE,3.06) (OK,30.66) (SC,4.50) (WA,10.95) (WI,4.99)};
\addplot[ybar, fill=qr5!50, draw=qr5] plot coordinates {(CA,34.00) (CO,5.29) (GA,0.00) (KY,8.02) (MD,16.96) (MT,9.39) (NC,53.93) (NE,10.01) (OK,18.56) (SC,6.62) (WA,4.99) (WI,14.21)};
\legend{\strut 1, \strut 2, \strut 3, \strut 4, \strut 5}
\end{axis}
\end{tikzpicture}
    \vspace{-2em}
    \caption{Quality rating score distribution.}
    \label{fig:qr_dist}
    \end{subfigure}
    \hfill
    \centering
    \begin{subfigure}[b]{0.48\textwidth}
    \centering
    \begin{tikzpicture}[scale=0.95]
\begin{axis}[
  boxplot/every box/.style={draw=black!55, fill=gray!8},
  boxplot/every median/.style={draw=black!75, thick},
  boxplot/draw direction=y,
  xtick={1,2,3,4,5,6,7,8,9,10,11,12},
  xticklabels={CA, CO, GA, KY, MD, MT, NC, NE, OK, SC, WA, WI},
  x tick label style={yshift=-7pt, rotate=45, anchor=center, font=\scriptsize},
  y tick label style={font=\footnotesize},
  ylabel={Feature missing rate (\%)},
  ylabel near ticks,
  ylabel style={font=\small},
  ]
    \addplot[black!55, boxplot prepared={
        lower whisker=0.12, lower quartile=82.76,
        median=95.75, upper quartile=97.76,
        upper whisker=99.18},
    ] coordinates {};

    \addplot[black!55, boxplot prepared={
        lower whisker=0.00, lower quartile=0.04,
        median=39.12, upper quartile=99.07,
        upper whisker=99.85},
    ] coordinates {};

    \addplot[black!55, boxplot prepared={
        lower whisker=3.73, lower quartile=3.80,
        median=55.37, upper quartile=83.47,
        upper whisker=99.90},
    ] coordinates {};

    \addplot[black!55, boxplot prepared={
        lower whisker=0.00, lower quartile=0.42,
        median=38.66, upper quartile=61.37,
        upper whisker=98.53},
    ] coordinates {};

    \addplot[black!55, boxplot prepared={
        lower whisker=11.31, lower quartile=11.31,
        median=98.38, upper quartile=99.64,
        upper whisker=99.89},
    ] coordinates {};

    \addplot[black!55, boxplot prepared={
        lower whisker=0.00, lower quartile=6.35,
        median=88.95, upper quartile=93.92,
        upper whisker=96.49},
    ] coordinates {};

    \addplot[black!55, boxplot prepared={
        lower whisker=0.00, lower quartile=0.00,
        median=20.75, upper quartile=89.66,
        upper whisker=99.08},
    ] coordinates {};

    \addplot[black!55, boxplot prepared={
        lower whisker=0.00, lower quartile=5.75,
        median=5.93, upper quartile=13.42,
        upper whisker=76.97},
    ] coordinates {};

    \addplot[black!55, boxplot prepared={
        lower whisker=0.00, lower quartile=87.82,
        median=99.02, upper quartile=99.73,
        upper whisker=99.80},
    ] coordinates {};

    \addplot[black!55, boxplot prepared={
        lower whisker=0.00, lower quartile=0.00,
        median=0.00, upper quartile=0.00,
        upper whisker=3.77},
    ] coordinates {};

    \addplot[black!55, boxplot prepared={
        lower whisker=0.00, lower quartile=15.83,
        median=93.88, upper quartile=99.53,
        upper whisker=99.84},
    ] coordinates {};

    \addplot[black!55, boxplot prepared={
        lower whisker=12.43, lower quartile=88.97,
        median=94.53, upper quartile=95.65,
        upper whisker=99.55},
    ] coordinates {};

\end{axis}
\end{tikzpicture}
    \vspace{-2em}
    \caption{Feature missing rate.}
    \label{fig:missing_ratio}
    \end{subfigure}
    \caption{Overview of the CCQ dataset. (a) Distribution of quality rating scores across rated providers in each state. (b) Feature missing rates for each state's \textit{raw} release, excluding the provider identifier and rating column. Boxes span the interquartile range and whiskers mark the 5th and 95th percentiles.}
\label{fig:dataset_overview}
\end{figure}

Rating coverage and rating distributions also vary widely across states. The rated share of a state's providers runs from 7.9\% in CA to 99.9\% in OK (Table~\ref{tab:data_source}). Figure~\ref{fig:qr_dist} shows the share of rated providers at each level of a state's scale. Every state's rating distribution except OK's is skewed, but the direction of the skew is not shared. Five states concentrate at the bottom of their scale: KY and MD place 61.23\% and 43.60\% of providers at one star, and NE, SC, and CO place 76.55\%, 75.31\%, and 63.51\% of providers in their bottom two levels. Two states concentrate at the top: CA assigns four- and five-star ratings to 87.36\% of providers, and NC assigns five stars to 53.93\%. Four states concentrate in the middle: WA places 80.24\% of providers at exactly three stars, WI places 52.08\% at two stars, MT places 65.19\% at two or three stars, and GA places 65.86\% at two stars within its three-level scale. Only OK resembles a balanced distribution. In addition, some classes are extremely small: WA, WI, MT, and CA report 0, 2, 6, and 6 providers, respectively, at the one-star level.

Furthermore, the feature missing rate differs across states. Figure~\ref{fig:missing_ratio} summarizes, for each state, the distribution across that state's feature columns of the percentage of providers for which the column carries no value. The figure describes the \textit{raw} release, in which an attribute a provider does not report is left empty rather than encoded (Section~\ref{sssec:data_curation-clean}). On the lower end, SC is nearly complete: seven of its eight feature columns carry a value for every provider, and only capacity is ever missing (5.81\%). NE (5.93\%) and NC (20.75\%) sit below a third of empty values for the median field, followed by KY (38.66\%), CO (39.12\%), and GA (55.37\%). Lastly, MT (88.95\%), WA (93.88\%), WI (94.53\%), CA (95.75\%), MD (98.38\%), and OK (99.02\%) have nearly empty columns for their median field. MD shows the sharpest split of the twelve: only one of its 30 feature columns (program type) is fully populated, nine more are missing for at most 15\% of providers, and 19 of the remaining 20 are empty for more than 86\%.

\subsection{Dataset Release}
\label{sec:dataset-release}
To promote reproducibility and facilitate future research, the \textit{CCQ} dataset is made publicly available exclusively for non-commercial, academic research purposes. The dataset is hosted on the Hugging Face platform (\url{https://huggingface.co/datasets/GAIN-Lab/CCQ}) as a gated dataset. Access is granted upon request: prospective users are required to sign a data use agreement, specify the intended purpose of use, and acknowledge the non-commercial restriction. This mechanism ensures that all usage complies with the CC-BY-NC-SA 4.0 license and the ethical guidelines governing the dataset. The code for data curation and for reproducing the benchmarking experiments is available at \url{https://github.com/Veeeeeee7/CCQ}.
\section{Benchmark Design}
\label{sec:benchmark_design}

Using \textit{CCQ}, we design a benchmark centered on the QR prediction task: given the curated record of a single child care provider, predict the quality rating that its state would assign. The benchmark contains two tasks that mirror how state agencies would deploy such models. In within-state prediction, a model is trained on a state's own rated providers and predicts ratings for the rest. In cross-state prediction, a model for one state is trained on other states, whose schemas and rating systems differ. 

\subsection{Within-State QR Prediction}
\label{sec:task-within}
States use QRIS to evaluate the quality of child care providers and communicate it to families~\cite{jenkins2021participates}. However, most QRIS do not rate every provider~\cite{ecqa2019qris}. Participation is voluntary in most states~\cite{jenkins2021participates}, and ratings rely on in-person observations that are costly and time-consuming to collect~\cite{markowitz2020simplifying}. Accurate within-state QR prediction can help state agencies close this coverage gap by assigning approximate ratings to unrated providers. Such models could also lower the cost of rating and extend ratings to more providers.

Formally, let $\mathcal{D} = \{(\mathbf{x}_i, y_i)\}_{i=1}^{n}$ denote the dataset of a single state, where $\mathbf{x}_i \in \mathcal{X}$ is the feature vector of child care provider $i$ and $y_i \in \mathcal{Y}$ is its quality rating. The label space $\mathcal{Y} = \{1,\dots,K\}$ is ordinal. In the within-state setting, the training set $\mathcal{D}^{\text{train}}$ and the test set $\mathcal{D}^{\text{test}}$ are both drawn \textit{i.i.d.} from the same distribution $P$ over $\mathcal{X} \times \mathcal{Y}$. They therefore share the same feature schema $\mathcal{X}$ and label space $\mathcal{Y}$. The task is to learn a classifier $f: \mathcal{X} \to \mathcal{Y}$ from $\mathcal{D}^{\text{train}}$ that performs best on the held-out $\mathcal{D}^{\text{test}}$. Since there is no schema mismatch or distribution shift, this is a standard supervised ordinal classification problem.

The evaluation protocol should follow common practices in supervised machine learning. We conduct stratified five-fold cross-validation~\cite{kohavi1995cv}, with the fold assignments fixed once per state and shared by all compared methods. Since quality ratings are ordinal, we use ordinal-aware metrics such as quadratic weighted kappa (QWK)~\cite{cohen1968qwk}. Balanced accuracy (BA)~\cite{brodersen2010balanced} is also reported as a complementary measure that accounts for label imbalance. This matters for \textit{CCQ}, where some rating levels hold only a handful of providers (Section~\ref{sec:dataset-overview}).

\subsection{Cross-State QR Prediction}
\label{sec:task-cross}
In practice, QRIS coverage varies widely across states and depends on local policy. Among the QRIS that reported participation in 2017, the share of licensed centers with a rating ranged from 1\% in New Jersey to 100\% in states whose QRIS places every licensed program at the first level~\cite{ecqa2019qris}. High-coverage states could therefore lend their data to states whose rated subset is too small to train a within-state model. However, each QRIS uses its own data schema, variable definitions, and QR scale (e.g., 1--5 vs. 1--3). This creates a need for models that transfer knowledge across states despite limited and incompatible labeled data.

Formally, let $\mathcal{D}_s$ denote the pooled dataset of a set of source states, with feature space $\mathcal{X}_s$ and ordinal label space $\mathcal{Y}_s$. Let $\mathcal{D}_t$ denote the dataset of a target state $t$, with feature space $\mathcal{X}_t$ and label space $\mathcal{Y}_t$. In general, $\mathcal{X}_s \neq \mathcal{X}_t$ and $\mathcal{Y}_s \neq \mathcal{Y}_t$. The marginal distributions $P_s(\mathbf{x})$ and $P_t(\mathbf{x})$ may also differ, as may the conditional distributions $P_s(y|\mathbf{x})$ and $P_t(y|\mathbf{x})$. The task is to learn a model $f: \mathcal{X}_t \to \mathcal{Y}_t$ from the labeled source data $\mathcal{D}_s^{\text{train}}$, with some labeled target data, so that it generalizes to the held-out target test set $\mathcal{D}_t^{\text{test}}$. A model must therefore handle feature schema mismatch, label space mismatch, and distribution shift at the same time.

We adopt a \textit{leave-one-state-out} protocol for evaluating cross-state QR prediction. It is the state-level analogue of holding out one domain for testing in domain generalization~\cite{gulrajani2021dg}. Each state $t$ serves as the target once, and the remaining states $\{\mathcal{D}_{s_1}, \mathcal{D}_{s_2}, \dots\}$ are pooled into a single source training set. To reflect real-world conditions, we evaluate both zero-shot and target-supervised transfer settings. First, a stratified 20\% test set is held out from the target state $\mathcal{D}_t$ and used to score every setting. Holding this test set fixed keeps scores comparable across supervision levels. Then, a stratified draw of $p\%$ of the target state's providers is created as the adaptation set, with $0 \leq p \leq 80$. At $p{=}0$, the setting is zero-shot, and no target label is used anywhere in training. At $p{=}80$, the adaptation set is the entire 80\% that remains after the test set is held out. For levels in between, the providers outside the draw are left unused. At every level, all source states provider records are used for training. We evaluate cross-state prediction with the same metrics as within-state prediction (QWK and BA).

\section{Experiments}
\label{sec:experiments}

\subsection{Experimental Setup}
\label{sec:exp-setup}
\subsubsection{Compared Models}
\label{sssec:models}
We evaluate two groups of models on \textit{CCQ} for both within- and cross-state QR prediction tasks. The first group, \textbf{tabular classifiers}, consumes the \textit{preprocessed} release natively or the \textit{raw} release embedded as a numerical table. The second group, \textbf{textual classifiers}, consumes the \textit{raw} release by processing text natively. Each tabular classifier is marked with the representation it consumes: \textit{-num} for \textit{preprocessed} and \textit{-txt} for \textit{raw}. The textual classifiers consume only the \textit{raw} representation and carry no suffix. The compared models are:

\begin{itemize}
    \item \textbf{Tabular Classifiers}: (a.1) a majority-class baseline derived from the training set (\textit{Dummy}); (a.2) logistic regression (\textit{LR}); (a.3) random forest (\textit{RF})~\cite{breiman2001rf}; (a.4) XGBoost (\textit{XGB})~\cite{chen2016xgboost}; (a.5) a tabular neural network (\textit{TabNN})~\cite{arik2021tabnet}; and (a.6) a tabular prior-fitted network (\textit{TabPFN})~\cite{hollmann2025tabpfn}.
    \item \textbf{Textual Classifiers}: (b.1) ModernBERT~\cite{warner2025modernbert} finetuned on textualized records with a classification head (\textit{MBERT}); (b.2) the same model trained with the CORN rank-consistent ordinal loss~\cite{shi2023corn} instead of cross-entropy (\textit{MBERT-CORN}); (b.3) an LLM (Qwen3-4B~\cite{yang2025qwen3}) with a finetuned classification head over the frozen backbone (\textit{LLM-CLS}); (b.4) the same frozen-backbone configuration with the most representative labeled provider of each class prepended to the input, a simplified form of retrieval augmentation~\cite{lewis2020rag} (\textit{LLM-RAG}); and (b.5) the same LLM with LoRA adapters~\cite{hu2022lora} trained alongside the classification head (\textit{LLM-LoRA}).
\end{itemize}

For cross-state QR prediction (Section~\ref{sec:task-cross}), we evaluate six target supervision levels, $p \in \{0, 16, 32, 48, 64, 80\}$. The models follow one of two training regimes. Dummy, LR, RF, XGB, and TabPFN are one-shot-fit models, which are fit once on the union of the pooled source states and the target draw. TabNN, MBERT, MBERT-CORN, and the three LLM variants are sequentially trained models, which are first pretrained on the pooled source states and then finetuned on the target draw.

\subsubsection{Implementation Details}
\label{sssec:implementation}
Table~\ref{tab:hparams} lists the complete configuration of every model. We use scikit-learn~\cite{pedregosa2011sklearn} for Dummy, LR, and RF, PyTorch~\cite{paszke2019pytorch} for the neural models, and Hugging Face Transformers~\cite{wolf2020transformers} for the language models. We tune hyperparameters only when required for convergence or for consistency across experiments, and otherwise run every method at its library defaults. Our goal is a uniform set of baselines rather than a per-experiment optimum, and a single configuration keeps comparisons consistent and fair across states. Since class imbalance is consistently present in \textit{CCQ}, we apply class weights, or subsampling when a model has no weighting mechanism.

\begin{table*}[!htbp]
\centering
\footnotesize
\setlength{\tabcolsep}{4pt}
\renewcommand{\arraystretch}{1.15}
\caption{Complete hyperparameter configuration. Values are held fixed across all 12 states, both tasks, and all six supervision levels. ``Phase'' is either (1) source pretraining or (2) target adaptation in sequential training. Each phase derives its own class weights and validation split.}
\label{tab:hparams}
\begin{tabular}{@{}l p{0.85\textwidth}@{}}
\toprule
\textbf{Component} & \textbf{Configuration} \\
\midrule
\multicolumn{2}{@{}l}{\textit{Training Protocol}} \\
Seed & 42 globally (splits, folds, subsamples, initializations, bootstrap) \\
Cross-validation & 5-fold stratified, shuffled; folds cached per state and shared by all methods \\
Target test block & stratified 20\%, carved once per target, scored at every supervision level \\
Validation split & stratified 15\% of each training phase; best epoch by validation QWK, patience 2 (textual classifiers; TabNN below) \\
\midrule
\multicolumn{2}{@{}l}{\textit{Tabular classifiers}} \\
Dummy & most-frequent class \\
LR & $\ell_2$ penalty, $C{=}1.0$, lbfgs, max 2,000 iterations, balanced class weights \\
RF & 100 trees, unlimited depth, 1 sample per leaf, $\sqrt{d}$ features per split, balanced subsample weights \\
XGB & 500 rounds, depth 6, learning rate 0.05, row subsample 0.8, column subsample 0.8, $\lambda{=}1.0$, histogram tree method, per-row balanced sample weights \\
TabNN & $n_d{=}n_a{=}16$, 3 steps, $\gamma{=}1.5$, $\lambda_{\text{sparse}}{=}10^{-4}$, entmax masks~\cite{peters2019entmax}; Adam~\cite{kingma2015adam} at $2{\times}10^{-2}$ with StepLR($20$, $0.9$); $\leq$200 epochs, patience 40 on balanced accuracy; batch $\min(256, n_{\text{train}})$ with virtual batch $\min(64,\text{batch})$; inverse-frequency class weights; transfer phase 2 warm-starts from phase 1 \\
TabPFN & TabPFN v2, 8 ensemble members, in-context fitting (no gradient training); training set class-balanced by downsampling to the smallest class, then capped at 10,000 context rows \\
\midrule
\multicolumn{2}{@{}l}{\textit{Textual classifiers}} \\
MBERT & ModernBERT-base, 8,192-token window, classification head; class-weighted cross-entropy, AdamW~\cite{loshchilov2019adamw} at $2{\times}10^{-5}$, linear decay with 10\% warmup, weight decay 0.01, 4 epochs per phase, batch 8, length-grouped batches with dynamic padding \\
MBERT-CORN & identical backbone and optimization trained with the CORN rank-consistent ordinal loss, class-weighted by the same balanced weights as MBERT; prediction by the canonical cumulative-sigmoid count decode, probabilities by telescoping the cumulative terms \\
LLM-CLS & Qwen3-4B (dense) with the backbone frozen and only a classification head trained; head learning rate $10^{-3}$, 4 epochs per phase, batch 8 with gradient accumulation 2 (effective 16), 8,192-token window, 10\% warmup, weight decay 0.01, balanced class weights \\
LLM-RAG & identical head and optimization; each input is additionally prefixed with one exemplar per rating level, chosen as the MiniLM class medoid. Exemplars are drawn from the training pool only and self-excluded, so no scored provider ever supplies or receives its own exemplar; in the zero-shot setting exemplars come from the source pool, since no target labels exist \\
LLM-LoRA & same backbone with LoRA adapters ($r{=}16$, $\alpha{=}32$, dropout 0.05) on the attention and MLP projections, learning rate $10^{-4}$, gradient checkpointing enabled \\
\bottomrule
\end{tabular}

\end{table*}

\begin{figure}[!htbp]
    \centering
    \resizebox{\linewidth}{!}{\input{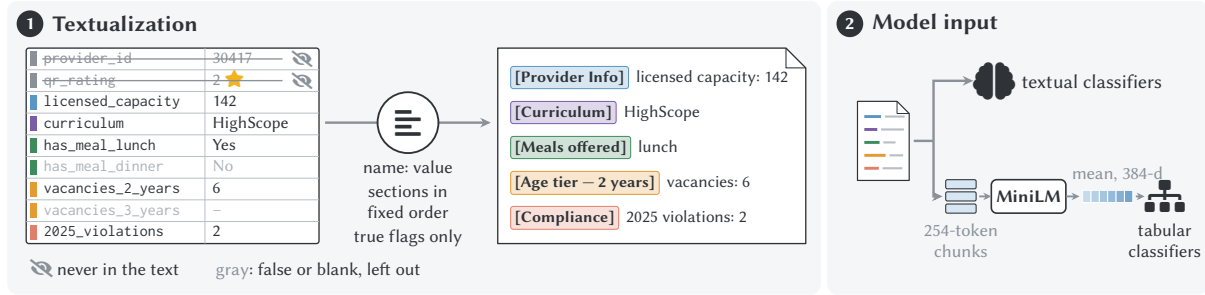}%
\begin{tikzpicture}[
  x=1cm, y=1cm,
  font=\sffamily\fontsize{7.4}{8.6}\selectfont,
  every node/.style={inner sep=0pt, outer sep=0pt, text=ccqInk},
  stage/.style={circle, fill=ccqInk, text=white, minimum size=10pt,
                font=\sffamily\bfseries\fontsize{6.8}{7}\selectfont},
  ttl/.style={anchor=west, font=\sffamily\bfseries\fontsize{8.4}{9}\selectfont},
  lbl/.style={align=center, font=\sffamily\fontsize{7.4}{8.4}\selectfont},
  note/.style={align=center, text=ccqMute, font=\sffamily\fontsize{6.8}{7.6}\selectfont},
  fld/.style={anchor=base west, font=\ttfamily\fontsize{6.0}{6.0}\selectfont},
  val/.style={anchor=base west, font=\sffamily\fontsize{6.4}{6.4}\selectfont},
  chip/.style={anchor=base west, fill=#1!24, draw=#1, line width=0.35pt, rounded corners=1pt,
               inner xsep=1.4pt, inner ysep=1.5pt, font=\sffamily\bfseries\fontsize{6.0}{6.0}\selectfont},
  txt/.style={anchor=base west, font=\sffamily\fontsize{6.4}{6.4}\selectfont},
  flow/.style={-{Stealth[length=4pt,width=3.6pt]}, draw=ccqMute, line width=0.9pt},
  thin/.style={-{Stealth[length=3pt,width=2.8pt]}, draw=ccqMute, line width=0.7pt},
]
\node[stage] (n1) at (0.30,-0.30) {1};
\node[ttl] at ($(n1.east)+(0.12,0)$) {Textualization};

\def\ccqRowH{0.285}
\draw[ccqInk, fill=white, line width=0.5pt, rounded corners=1pt] (0.25,-0.62) rectangle (4.10,-3.19);
\foreach \i in {1,...,8}{\draw[ccqLine, line width=0.3pt] (0.25,-0.62-\i*\ccqRowH) -- (4.10,-0.62-\i*\ccqRowH);}
\draw[ccqLine, line width=0.3pt] (2.60,-0.62) -- (2.60,-3.19);
\def\ccqRow#1#2#3#4#5{%
  \fill[#2] (0.31,{-0.62-(#1-1)*\ccqRowH-0.05}) rectangle (0.39,{-0.62-#1*\ccqRowH+0.05});
  \node[fld, text=#5] at (0.48,{-0.62-#1*\ccqRowH+0.09}) {#3};
  \node[val, text=#5] at (2.70,{-0.62-#1*\ccqRowH+0.09}) {#4};}
\ccqRow{1}{ccqMute}{provider\_id}{30417}{ccqMute}
\ccqRow{2}{ccqMute}{qr\_rating}{2 \textcolor{ccqStar}{\faIcon{star}}}{ccqMute}
\ccqRow{3}{ccqData}{licensed\_capacity}{142}{ccqInk}
\ccqRow{4}{ccqProp}{curriculum}{HighScope}{ccqInk}
\ccqRow{5}{ccqOpen}{has\_meal\_lunch}{Yes}{ccqInk}
\ccqRow{6}{ccqOpen}{has\_meal\_dinner}{No}{ccqMute!70}
\ccqRow{7}{ccqTgt}{vacancies\_2\_years}{6}{ccqInk}
\ccqRow{8}{ccqTgt}{vacancies\_3\_years}{\textendash}{ccqMute!70}
\ccqRow{9}{ccqPos}{2025\_violations}{2}{ccqInk}
\foreach \i in {1,2}{%
  \draw[ccqMute, line width=0.5pt] (0.46,{-0.62-(\i-0.5)*\ccqRowH}) -- (3.62,{-0.62-(\i-0.5)*\ccqRowH});
  \node[text=ccqMute, font=\fontsize{6.2}{6.2}\selectfont] at (3.88,{-0.62-(\i-0.5)*\ccqRowH}) {\faIcon{eye-slash}};}

\draw[flow] (4.18,-1.60) -- (6.36,-1.60);
\node[station, font=\fontsize{11}{11}\selectfont] (tx) at (5.27,-1.60) {\faIcon{align-left}};
\node[note, text=ccqInk, anchor=north] at (5.27,-2.10)
  {name: value\\[1.5pt] sections in\\ fixed order\\[1.5pt] true flags only};

\draw[ccqInk, fill=white, line width=0.5pt, line join=round]
  (6.45,-3.19) -- (6.45,-0.62) -- (10.24,-0.62) -- (10.50,-0.88) -- (10.50,-3.19) -- cycle;
\draw[ccqInk, line width=0.4pt, line join=round] (10.24,-0.62) -- (10.24,-0.88) -- (10.50,-0.88);
\foreach \y/\c/\h/\t in {-1.06/ccqData/{[Provider Info]}/{licensed capacity: 142},
                         -1.52/ccqProp/{[Curriculum]}/{HighScope},
                         -1.98/ccqOpen/{[Meals offered]}/{lunch},
                         -2.44/ccqTgt/{[Age tier \textemdash{} 2 years]}/{vacancies: 6},
                         -2.90/ccqPos/{[Compliance]}/{2025 violations: 2}}{%
  \node[chip=\c] (h) at (6.62,\y) {\h};
  \node[txt] at ($(h.base east)+(0.10,0)$) {\t};}

\node[anchor=west, note, text=ccqInk] at (0.30,-3.53)
  {\textcolor{ccqMute}{\faIcon{eye-slash}}\ never in the text\qquad
   \textcolor{ccqMute}{gray}: false or blank, left out};

\node[stage] (n2) at (11.08,-0.30) {2};
\node[ttl] at ($(n2.east)+(0.12,0)$) {Model input};

\draw[ccqInk, fill=white, line width=0.45pt, line join=round]
  (11.18,-2.36) -- (11.18,-1.32) -- (11.70,-1.32) -- (11.86,-1.48) -- (11.86,-2.36) -- cycle;
\draw[ccqInk, line width=0.35pt, line join=round] (11.70,-1.32) -- (11.70,-1.48) -- (11.86,-1.48);
\foreach \y/\c/\a/\b in {-1.52/ccqData/0.22/0.26, -1.69/ccqProp/0.17/0.30, -1.86/ccqOpen/0.20/0.16,
                         -2.03/ccqTgt/0.28/0.18, -2.20/ccqPos/0.19/0.26}{%
  \draw[\c, line width=1.1pt] (11.27,\y) -- ++(\a,0);
  \draw[ccqLine, line width=0.8pt] ({11.27+\a+0.05},\y) -- ++(\b,0);}
\draw[ccqMute, line width=0.9pt] (11.92,-1.84) -- (12.18,-1.84);
\node[font=\fontsize{14}{14}\selectfont] (brain) at (12.98,-1.05) {\faIcon{brain}};
\node[lbl, anchor=west] at (13.34,-1.05) {textual classifiers};
\draw[flow] (12.18,-1.84) |- (brain.west);
\foreach \k in {0,1,2}{%
  \draw[ccqInk, fill=ccqDataLt, line width=0.4pt, rounded corners=0.6pt]
    (12.34,{-2.47-\k*0.16}) rectangle (12.74,{-2.35-\k*0.16});}
\draw[flow] (12.18,-1.84) |- (12.30,-2.57);
\node[note, anchor=north] at (12.54,-2.93) {254-token\\chunks};
\node[draw=ccqInk, fill=white, line width=0.6pt, rounded corners=2pt, minimum width=0.98cm,
      minimum height=0.44cm, font=\sffamily\bfseries\fontsize{6.6}{6.6}\selectfont] (lm) at (13.43,-2.57) {MiniLM};
\draw[thin] (12.78,-2.57) -- (lm.west);
\foreach \k/\t in {0/35, 1/45, 2/55, 3/65, 4/75, 5/85}{%
  \fill[ccqData!\t, draw=white, line width=0.3pt] ({14.14+\k*0.11},-2.49) rectangle ({14.25+\k*0.11},-2.65);}
\draw[thin] (lm.east) -- (14.10,-2.57);
\node[note, anchor=south] at (14.60,-2.37) {mean, 384-d};
\node[font=\fontsize{12}{12}\selectfont] (tree) at (15.22,-2.57) {\faIcon{sitemap}};
\draw[thin] (14.84,-2.57) -- (tree.west);
\node[note, text=ccqInk, anchor=north] at (15.22,-2.93) {tabular\\classifiers};

\begin{scope}[on background layer]
  \fill[ccqZone, rounded corners=4pt] (0,0) rectangle (10.70,-3.86);
  \fill[ccqZone, rounded corners=4pt] (10.78,0) rectangle (15.80,-3.86);
\end{scope}
\end{tikzpicture}%
}
    \caption{Text-based representation of a provider record, following PedCA-FT~\cite{lu2025PedCAFT}. \textbf{(1)} A simplified record in Georgia's schema is written as text. False and blank values are left out, and the provider identifier and the rating never appear. \textbf{(2)} The textual classifiers read this text directly. For the tabular classifiers, all-MiniLM-L6-v2 embeds the text in 254-token chunks, and the normalized mean of the chunk embeddings is the 384-dimensional input.}
    \label{fig:text_representation}
\end{figure}

The tabular classifiers (a.1--a.6) natively consume the \textit{preprocessed} representation (\textit{-num}), while the textual classifiers (b.1--b.5) consume the \textit{raw} representation. To write each \textit{raw} record as text, we follow the text-based representation of PedCA-FT~\cite{lu2025PedCAFT} (Figure~\ref{fig:text_representation}). Each record becomes natural-language text, with its fields grouped into labeled sections in a fixed order: general provider information, curriculum, facility capabilities, enrollment details per age group, and multi-year compliance history. Boolean fields list only their true values and missing values are omitted. For within-state QR prediction, the tabular classifiers are evaluated on both representations. On the \textit{raw} representation, each textualized record is embedded with the sentence encoder \textit{all-MiniLM-L6-v2}~\cite{reimers2019sentencebert,wang2020minilm}. Records longer than the encoder window are split into chunks, and the unit-normalized chunk embeddings are mean-pooled and normalized, so no record is truncated. The textual classifiers are evaluated on the textualized records only. In cross-state QR prediction, all models use the \textit{raw} representation, since the \textit{preprocessed} one follows state-specific schemas that cannot be pooled.

\subsubsection{Evaluation Metrics}
\label{sec:metrics}
For both within- and cross-state QR prediction, we report balanced accuracy (BA)~\cite{brodersen2010balanced} and quadratic weighted kappa (QWK)~\cite{cohen1968qwk} (Section~\ref{sec:benchmark_design}). Since quality ratings are ordinal, QWK is the primary metric. It penalizes each error by its squared distance on the rating scale and corrects for the agreement expected by chance. Within-state results are the mean over the five folds, and cross-state results are a single point estimate on the held-out target test set.

To summarize overall performance across states and metrics, we also report Elo scores~\cite{boubdir2024elo}. Elo gives each model a single rating derived from pairwise comparisons. Each combination of state and metric is one match between every pair of models, and the model with the higher score wins. Following existing work~\cite{boubdir2024elo,erickson2025tabarena}, ratings are updated after each match, with a gap of $\Delta$ points giving the higher-rated model an expected win rate of $1/(1+10^{-\Delta/400})$. A 400-point gap therefore corresponds to a 91\% expected win rate, and equal ratings to 50\%. As in TabArena~\cite{erickson2025tabarena}, we anchor RF-txt at 1,000 points in both settings and report 95\% bootstrap confidence intervals. Each bootstrap round resamples the twelve states with replacement, and the reported score is the mean over 2,000 rounds. In the cross-state setting, the six supervision levels of a state are not independent samples. We therefore compute Elo separately at each level on the same resampled states and average it over the six levels, so that each level carries equal weight. Because the within- and cross-state ratings are computed from different sets of matches, scores can be compared within a setting but not across settings.

\subsection{Within-State QR Prediction Experiment}
\label{sec:results-within}
Table~\ref{tab:within} reports the within-state QR prediction results. Each state's models are trained and evaluated only on its own providers, using stratified five-fold cross-validation on that state's native rating scale. Tabular classifiers are evaluated on both representations, and textual classifiers on the \textit{raw} representation only. Figure~\ref{fig:elo_within_state} summarizes overall performance with Elo scores computed from BA and QWK across all twelve states. Figure~\ref{fig:pareto_within_state} shows the trade-off between performance and model complexity as the Pareto frontier of Elo score versus the number of tuned parameters.

\begin{table}[!htbp]
\centering
\scriptsize
\setlength{\tabcolsep}{3pt}
\caption{Within-state QR prediction results on each state's native QR scale. Each cell is the mean over five-fold cross-validation and reports BA and QWK, both as percentages. The best value per state and metric is in \textbf{bold}, and the second best is \underline{underlined}.}
\label{tab:within}
\resizebox{\textwidth}{!}{%
\begin{tabular}{l|cc|cc|cc|cc|cc|cc|cc|cc|cc|cc|cc|cc}
\toprule
\multirow{2}{*}{\textbf{Method}} & \multicolumn{2}{c|}{\textbf{CA}} & \multicolumn{2}{c|}{\textbf{CO}} & \multicolumn{2}{c|}{\textbf{GA}} & \multicolumn{2}{c|}{\textbf{KY}} & \multicolumn{2}{c|}{\textbf{MD}} & \multicolumn{2}{c|}{\textbf{MT}} & \multicolumn{2}{c|}{\textbf{NC}} & \multicolumn{2}{c|}{\textbf{NE}} & \multicolumn{2}{c|}{\textbf{OK}} & \multicolumn{2}{c|}{\textbf{SC}} & \multicolumn{2}{c|}{\textbf{WA}} & \multicolumn{2}{c}{\textbf{WI}}\\
  & BA & QWK & BA & QWK & BA & QWK & BA & QWK & BA & QWK & BA & QWK & BA & QWK & BA & QWK & BA & QWK & BA & QWK & BA & QWK & BA & QWK\\
\midrule
Dummy & 20.00 & 0.00 & 20.00 & 0.00 & 33.33 & 0.00 & 20.00 & 0.00 & 20.00 & 0.00 & 20.00 & 0.00 & 20.00 & 0.00 & 20.00 & 0.00 & 20.00 & 0.00 & 20.00 & 0.00 & 25.00 & 0.00 & 23.00 & 0.00\\
\midrule
\multicolumn{12}{l}{\textit{Preprocessed feature representation}}\\
LR-num & \textbf{39.45} & 29.33 & \textbf{41.37} & 51.69 & \underline{45.23} & \underline{21.04} & 41.51 & 32.43 & 44.75 & 64.31 & 37.86 & 27.23 & \underline{50.29} & 44.90 & \underline{42.44} & \underline{29.03} & 49.48 & 64.64 & \underline{41.68} & 36.25 & \underline{43.51} & 27.09 & \textbf{44.71} & 50.59\\
RF-num & 34.01 & \textbf{44.82} & 37.06 & \underline{56.82} & 35.47 & 6.45 & 35.95 & 40.38 & 43.23 & 65.07 & \underline{42.67} & \textbf{31.67} & 42.37 & \underline{59.19} & 39.27 & 26.79 & 52.83 & 76.43 & 38.76 & 39.65 & 32.81 & 20.44 & 38.52 & \underline{52.75}\\
XGB-num & 36.22 & \underline{42.59} & 39.53 & \textbf{57.98} & 41.15 & 19.00 & \underline{41.76} & \textbf{48.10} & 44.27 & 65.23 & \textbf{43.85} & 26.63 & 46.02 & \textbf{61.73} & 38.84 & \textbf{29.99} & \underline{53.07} & \underline{76.44} & 39.66 & \underline{42.18} & 39.07 & \textbf{33.63} & 40.80 & 52.63\\
TabPFN-num & \underline{36.65} & 21.16 & \underline{40.67} & 51.24 & \textbf{48.04} & \textbf{23.33} & \textbf{45.93} & \underline{40.70} & \textbf{45.45} & 64.14 & 40.14 & \underline{27.49} & \textbf{54.15} & 51.18 & \textbf{46.22} & 26.74 & \textbf{56.43} & \textbf{77.34} & \textbf{42.76} & 38.66 & \textbf{44.32} & 24.88 & 30.74 & 18.17\\
TabNN-num & 24.78 & 18.58 & 36.22 & 46.92 & 40.18 & 14.27 & 38.13 & 29.53 & 44.90 & 63.82 & 39.85 & 20.62 & 48.40 & 48.01 & 42.09 & 18.57 & 48.76 & 64.75 & 40.28 & 37.72 & 40.93 & 21.37 & \underline{43.23} & 47.82\\
\midrule
\multicolumn{12}{l}{\textit{Raw feature representation}}\\
LR-txt & 33.66 & 29.80 & 35.66 & 42.34 & 39.36 & 10.85 & 38.90 & 33.18 & 44.18 & 64.38 & 40.14 & 26.53 & 48.27 & 49.81 & 38.22 & 26.16 & 43.95 & 62.58 & 40.32 & 37.98 & 35.58 & 20.39 & 37.70 & 37.86\\
RF-txt & 28.66 & 28.45 & 31.83 & 43.48 & 33.81 & 1.36 & 26.26 & 18.11 & 43.71 & 65.26 & 38.61 & 26.63 & 33.11 & 44.08 & 34.86 & 18.82 & 43.14 & 66.27 & 37.85 & 41.60 & 27.27 & 4.73 & 31.30 & 31.03\\
XGB-txt & 32.46 & 33.12 & 34.38 & 44.33 & 34.56 & 4.40 & 31.65 & 33.98 & 43.59 & 63.82 & 38.64 & 20.88 & 35.86 & 48.82 & 37.35 & 26.69 & 44.90 & 66.31 & 38.00 & \textbf{42.42} & 32.82 & 19.72 & 36.24 & 43.75\\
TabPFN-txt & 33.67 & 14.18 & 35.04 & 38.76 & 41.86 & 13.60 & 35.96 & 27.47 & \underline{45.36} & 64.35 & 37.56 & 10.62 & 43.16 & 43.42 & 36.07 & 13.34 & 49.09 & 65.61 & 41.35 & 40.80 & 41.79 & 19.92 & 19.80 & 6.10\\
TabNN-txt & 29.64 & 14.36 & 31.25 & 31.16 & 38.19 & 8.16 & 30.24 & 15.82 & 44.43 & 63.40 & 32.13 & 12.22 & 39.20 & 37.70 & 34.66 & 12.87 & 43.08 & 59.80 & 40.37 & 39.04 & 35.35 & 15.57 & 34.96 & 30.15\\
MBERT & 33.94 & 38.68 & 35.34 & 50.59 & 41.84 & 18.67 & 39.77 & 38.66 & 44.06 & \underline{70.63} & 23.57 & 12.35 & 38.38 & 48.66 & 36.49 & 19.58 & 51.58 & 70.80 & 40.11 & 42.02 & 38.11 & \underline{33.35} & 38.75 & 49.70\\
MBERT-CORN & 28.54 & 27.34 & 35.11 & 55.21 & 34.88 & 4.69 & 32.05 & 34.84 & 40.06 & \textbf{70.77} & 21.06 & 0.78 & 32.19 & 50.62 & 31.80 & 19.04 & 50.92 & 75.55 & 35.89 & 40.95 & 36.16 & 29.57 & 40.53 & \textbf{53.21}\\
LLM-CLS & 30.66 & 24.44 & 34.58 & 42.18 & 38.63 & 9.63 & 30.90 & 29.77 & 41.36 & 66.37 & 32.72 & 17.47 & 33.46 & 44.81 & 36.66 & 21.54 & 37.96 & 43.92 & 34.61 & 22.58 & 33.85 & 25.27 & 32.64 & 40.84\\
LLM-RAG & 24.10 & 11.75 & 33.52 & 43.32 & 39.23 & 10.32 & 32.57 & 26.80 & 41.96 & 66.08 & 28.31 & 8.55 & 32.85 & 41.28 & 33.50 & 18.74 & 36.31 & 42.69 & 38.74 & 35.09 & 35.13 & 27.00 & 29.25 & 30.88\\
LLM-LoRA & 28.02 & 25.58 & 35.18 & 46.96 & 42.74 & 17.49 & 35.35 & 34.88 & 42.52 & 68.80 & 35.71 & 23.67 & 36.72 & 47.48 & 38.64 & 24.95 & 50.07 & 68.92 & 40.84 & 38.01 & 35.02 & 27.09 & 34.50 & 50.42\\
\bottomrule
\end{tabular}}

\end{table}

\subsubsection{Predictive Performance}
\label{sssec:within-overall}
\begin{figure}[!htbp]
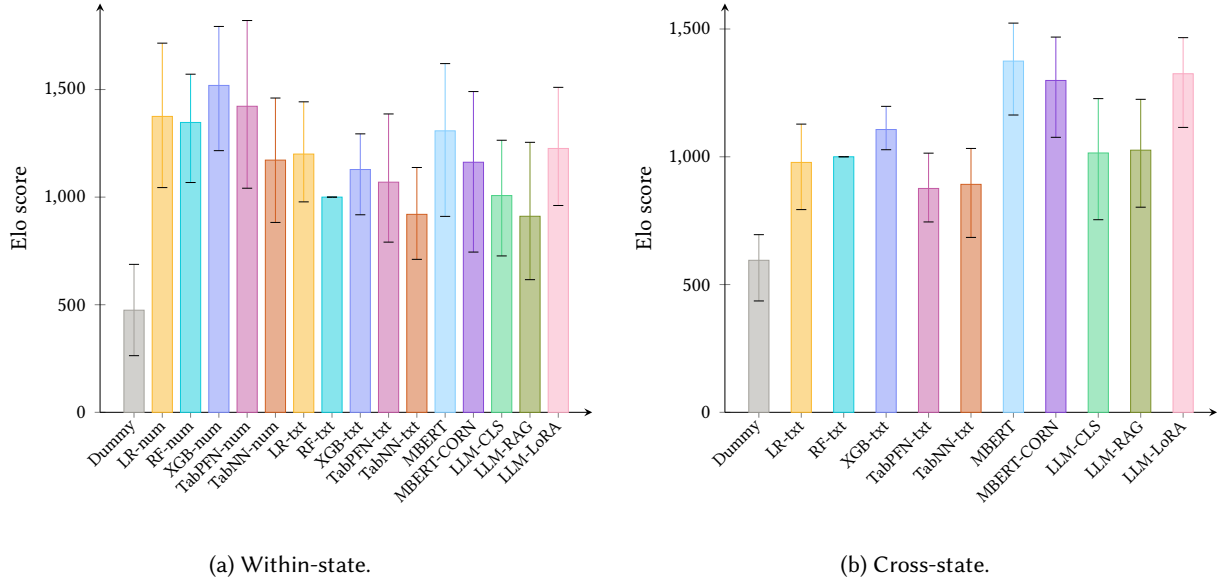

    \centering
    \begin{subfigure}[b]{0.48\textwidth}
        \centering
        \input{figs/model_colors}
\begin{tikzpicture}[scale=0.95]
\begin{axis}[
    ybar,
    axis lines=left,
    symbolic x coords={Dummy,LR-num,RF-num,XGB-num,TabPFN-num,TabNN-num,{LR-txt},{RF-txt},{XGB-txt},{TabPFN-txt},{TabNN-txt},{MBERT},{MBERT-CORN},{LLM-CLS},{LLM-RAG},{LLM-LoRA}},
    xtick={Dummy,LR-num,RF-num,XGB-num,TabPFN-num,TabNN-num,{LR-txt},{RF-txt},{XGB-txt},{TabPFN-txt},{TabNN-txt},{MBERT},{MBERT-CORN},{LLM-CLS},{LLM-RAG},{LLM-LoRA}},,
    x tick label style={yshift=-2pt, rotate=45, anchor=east, inner sep=0pt, font=\scriptsize},
    ylabel={Elo score},
    ylabel near ticks,
    label style={font=\small},
    y tick label style={font=\footnotesize},
    ymin=0, ymax=1900,
    bar width=8pt,
    enlarge x limits=0.08,
    error bars/y dir=both,
    error bars/y explicit,
    error bars/.style={gray,opacity=0.4},
    legend style={at={(0.5,1.05)}, anchor=south, legend columns=-1},
]

\addplot[ybar, bar shift=0pt, fill=cDummy!50, draw=cDummy] coordinates {(Dummy, 474.578) += (0, 212.963) -= (0, 211.299)};
\addplot[ybar, bar shift=0pt, fill=cLR!50, draw=cLR] coordinates {(LR-num, 1375.066) += (0, 340.468) -= (0, 330.698)};
\addplot[ybar, bar shift=0pt, fill=cRF!50, draw=cRF] coordinates {(RF-num, 1346.950) += (0, 224.295) -= (0, 279.008)};
\addplot[ybar, bar shift=0pt, fill=cXGB!50, draw=cXGB] coordinates {(XGB-num, 1519.051) += (0, 274.085) -= (0, 303.062)};
\addplot[ybar, bar shift=0pt, fill=cTabPFN!50, draw=cTabPFN] coordinates {(TabPFN-num, 1422.205) += (0, 398.466) -= (0, 380.823)};
\addplot[ybar, bar shift=0pt, fill=cTabNN!50, draw=cTabNN] coordinates {(TabNN-num, 1172.006) += (0, 288.503) -= (0, 289.627)};
\addplot[ybar, bar shift=0pt, fill=cLR!50, draw=cLR] coordinates {(LR-txt, 1200.092) += (0, 242.790) -= (0, 222.107)};
\addplot[ybar, bar shift=0pt, fill=cRF!50, draw=cRF] coordinates {(RF-txt, 1000.000) += (0, 0.000) -= (0, 0.000)};
\addplot[ybar, bar shift=0pt, fill=cXGB!50, draw=cXGB] coordinates {(XGB-txt, 1128.353) += (0, 165.670) -= (0, 210.348)};
\addplot[ybar, bar shift=0pt, fill=cTabPFN!50, draw=cTabPFN] coordinates {(TabPFN-txt, 1069.456) += (0, 317.073) -= (0, 278.558)};
\addplot[ybar, bar shift=0pt, fill=cTabNN!50, draw=cTabNN] coordinates {(TabNN-txt, 919.962) += (0, 217.729) -= (0, 209.119)};
\addplot[ybar, bar shift=0pt, fill=cMBERT!50, draw=cMBERT] coordinates {(MBERT, 1308.048) += (0, 312.501) -= (0, 397.566)};
\addplot[ybar, bar shift=0pt, fill=cCORN!50, draw=cCORN] coordinates {(MBERT-CORN, 1162.103) += (0, 328.483) -= (0, 417.428)};
\addplot[ybar, bar shift=0pt, fill=cCLS!50, draw=cCLS] coordinates {(LLM-CLS, 1007.307) += (0, 257.009) -= (0, 280.514)};
\addplot[ybar, bar shift=0pt, fill=cRAG!50, draw=cRAG] coordinates {(LLM-RAG, 911.012) += (0, 343.672) -= (0, 294.535)};
\addplot[ybar, bar shift=0pt, fill=cLoRA!50, draw=cLoRA] coordinates {(LLM-LoRA, 1226.119) += (0, 284.035) -= (0, 265.009)};

\end{axis}
\end{tikzpicture}
        \caption{Within-state.}
        \label{fig:elo_within_state}
    \end{subfigure}
    \hfill
    \begin{subfigure}[b]{0.48\textwidth}
        \centering
        \input{figs/model_colors}
\begin{tikzpicture}[scale=0.95]
\begin{axis}[
    ybar,
    axis lines=left,
    symbolic x coords={Dummy,LR-txt,RF-txt,XGB-txt,TabPFN-txt,TabNN-txt,MBERT,MBERT-CORN,LLM-CLS,LLM-RAG,LLM-LoRA},
    xtick={Dummy,LR-txt,RF-txt,XGB-txt,TabPFN-txt,TabNN-txt,MBERT,MBERT-CORN,LLM-CLS,LLM-RAG,LLM-LoRA},
    x tick label style={yshift=-2pt, rotate=45, anchor=east, inner sep=0pt, font=\scriptsize},
    ylabel={Elo score},
    ylabel near ticks,
    label style={font=\small},
    y tick label style={font=\footnotesize},
    ymin=0, ymax=1600,
    bar width=8pt,
    enlarge x limits=0.08,
    error bars/y dir=both,
    error bars/y explicit,
    error bars/.style={gray,opacity=0.4},
    legend style={at={(0.5,1.05)}, anchor=south, legend columns=-1},
]

\addplot[ybar, bar shift=0pt, fill=cDummy!50, draw=cDummy] coordinates {(Dummy, 594.902) += (0, 100.328) -= (0, 159.072)};
\addplot[ybar, bar shift=0pt, fill=cLR!50, draw=cLR] coordinates {({LR-txt}, 977.905) += (0, 149.939) -= (0, 184.594)};
\addplot[ybar, bar shift=0pt, fill=cRF!50, draw=cRF] coordinates {({RF-txt}, 1000.000) += (0, 0.000) -= (0, 0.000)};
\addplot[ybar, bar shift=0pt, fill=cXGB!50, draw=cXGB] coordinates {({XGB-txt}, 1106.692) += (0, 90.620) -= (0, 79.054)};
\addplot[ybar, bar shift=0pt, fill=cTabPFN!50, draw=cTabPFN] coordinates {({TabPFN-txt}, 876.027) += (0, 138.138) -= (0, 131.025)};
\addplot[ybar, bar shift=0pt, fill=cTabNN!50, draw=cTabNN] coordinates {({TabNN-txt}, 892.069) += (0, 140.509) -= (0, 207.657)};
\addplot[ybar, bar shift=0pt, fill=cMBERT!50, draw=cMBERT] coordinates {({MBERT}, 1374.425) += (0, 148.640) -= (0, 210.973)};
\addplot[ybar, bar shift=0pt, fill=cCORN!50, draw=cCORN] coordinates {({MBERT-CORN}, 1298.604) += (0, 169.800) -= (0, 222.452)};
\addplot[ybar, bar shift=0pt, fill=cCLS!50, draw=cCLS] coordinates {({LLM-CLS}, 1014.861) += (0, 212.754) -= (0, 261.109)};
\addplot[ybar, bar shift=0pt, fill=cRAG!50, draw=cRAG] coordinates {({LLM-RAG}, 1025.815) += (0, 199.005) -= (0, 223.138)};
\addplot[ybar, bar shift=0pt, fill=cLoRA!50, draw=cLoRA] coordinates {({LLM-LoRA}, 1324.948) += (0, 141.221) -= (0, 210.075)};

\end{axis}
\end{tikzpicture}
        \caption{Cross-state.}
        \label{fig:elo_cross_state}
    \end{subfigure}
    \caption{Elo ratings from the within-state and cross-state experiments. The Elo scale uses a 400-point gap, which corresponds to a 91\% expected win rate. Error bars show 95\% confidence intervals from 2,000 bootstrap rounds that resample the twelve states. Cross-state scores are averaged over the six supervision levels. RF-txt is the 1,000-point anchor in both panels. The two panels are computed on different sets of comparisons, so scores are comparable within a panel but not across panels.}
    \label{fig:elo_comparison}
\end{figure}

Tabular classifiers on the \textit{preprocessed} representation achieve the highest Elo scores (Figure~\ref{fig:elo_within_state}). XGB-num ranks first with 1519.1, followed by TabPFN-num (1422.2), LR-num (1375.1), and RF-num (1346.9). Every tabular classifier also scores higher on the \textit{preprocessed} representation than on the \textit{raw} one. Across LR, RF, XGB, TabPFN, and TabNN, the \textit{-num} version exceeds the \textit{-txt} version by 303.5 Elo points on average, an 85.2\% expected win rate. The gap is smallest for LR (175.0 points) and largest for XGB (390.7 points). On the \textit{raw} representation, MBERT is the strongest classifier, with an Elo score of 1308.0. It ranks fifth overall, ahead of TabNN-num (1172.0). Its expected win rate is 22.9\% against XGB-num and 44.4\% against RF-num. LLM-LoRA (1226.1) and LR-txt (1200.1) follow. MBERT-CORN scores 1162.1, 145.9 points below MBERT. Its QWK is higher than MBERT's in five states (CO, MD, NC, OK, and WI), but its BA is lower in 11 of the 12. Lastly, XGB-txt (1128.4), TabPFN-txt (1069.5), LLM-CLS (1007.3), RF-txt (1000.0), TabNN-txt (920.0), and LLM-RAG (911.0) form the lower tier, and the Dummy baseline scores 474.6.

However, Elo combines both metrics, and the two do not always agree. TabPFN-num owes much of its second place to BA, where it is best in 8 of the 12 states. On QWK, the primary metric, its mean of 38.75 falls below that of XGB-num (46.34), RF-num (43.37), MBERT (41.14), LR-num (39.88), and LLM-LoRA (39.52). The QWK ranking is thus led by the two tree ensembles on the \textit{preprocessed} representation. Table~\ref{tab:within} shows the same pattern state by state. A tabular classifier on the \textit{preprocessed} representation has the best QWK in 9 of the 12 states and the best BA in all 12. Specifically, XGB-num has the best QWK in five states, and RF-num and TabPFN-num in two each. The three exceptions are MD and WI, where MBERT-CORN is best (70.77 and 53.21), and SC, where XGB-txt is best (42.42). The margins in MD and WI are small. MBERT-CORN leads MBERT by 0.14 points in MD and RF-num by 0.46 points in WI.

The best QWK also varies widely across states. In OK and MD, the strongest models exceed 70. TabPFN-num reaches 77.34 and XGB-num 76.44 in OK, and MBERT-CORN reaches 70.77 and MBERT 70.63 in MD. In CA, CO, KY, NC, SC, and WI, the best QWK lies between 42.42 (SC) and 61.73 (NC). In the remaining four states, even the best model stays below 35, with 33.63 in WA, 31.67 in MT, 29.99 in NE, and 23.33 in GA. Overall, the best QWK per state spans 54 points, and the best model changes from state to state.

\subsubsection{Predictive Performance Versus Model Complexity}
\label{sssec:within-pareto}
\begin{figure}[!htbp]
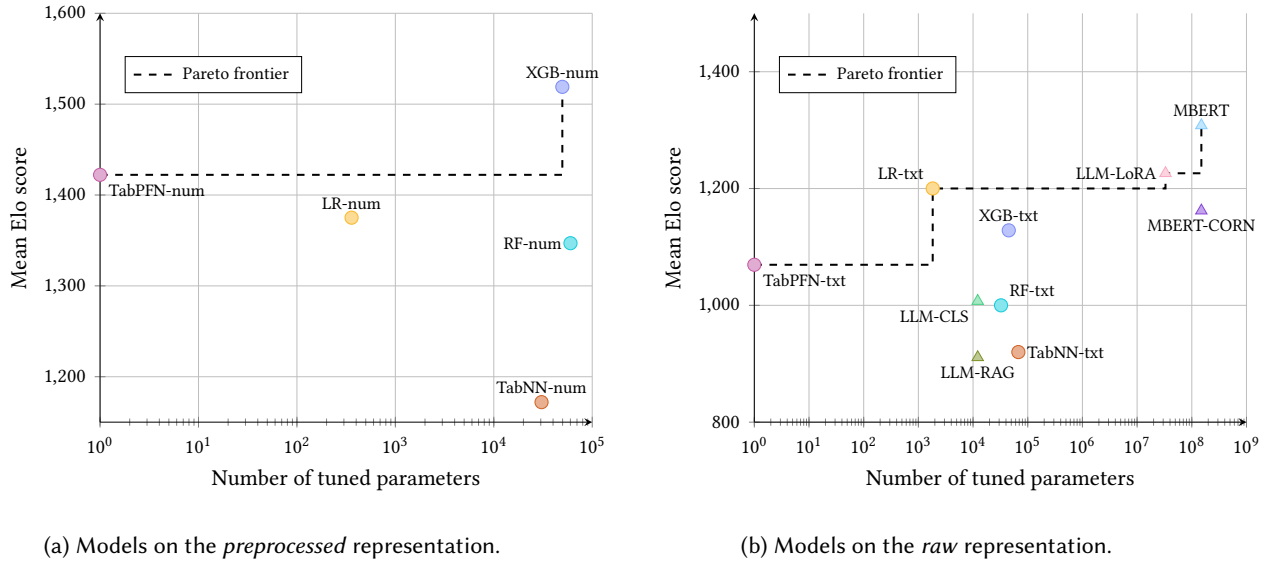

\centering
\begin{subfigure}[b]{0.45\textwidth}
    \centering
    \input{figs/model_colors}
\begin{tikzpicture}[scale=0.95]
\begin{axis}[%
axis lines=left,
xmin=0.0,
xmax=1e5,
ymax=1600,
ymin=1150,
xmajorgrids,
ymajorgrids,
legend style={at={(0.05,0.85)}, anchor=west, font=\scriptsize},
xlabel={Number of tuned parameters},
ylabel={Mean Elo score},
x tick label style={font=\footnotesize},
y tick label style={font=\footnotesize},
ylabel near ticks,
label style={font=\small},
xmode=log,
log basis x={10},
scatter/classes={%
    LR={mark=*,draw=cLR,fill=cLR!50},%
    RF={mark=*,draw=cRF,fill=cRF!50},%
    XGB={mark=*,draw=cXGB,fill=cXGB!50},%
    TabPFN={mark=*,draw=cTabPFN,fill=cTabPFN!50},%
    TabNN={mark=*,draw=cTabNN,fill=cTabNN!50}}]
\addplot[scatter,only marks,mark size=2.6pt,
nodes near coords style = {font=\scriptsize, text=black},
point meta=explicit symbolic,
nodes near coords*={\Method},
visualization depends on={value \thisrow{method} \as \Method},
visualization depends on={value \thisrow{anchor} \as \Anchor},
nodes near coords style = {anchor=\Anchor},
scatter src=explicit symbolic,
forget plot]%
table[meta=label] {
x       y               label   method      anchor
359      1375.066311     LR     LR-num      south
60130    1346.949584     RF     RF-num      east
49692    1519.050647     XGB    XGB-num     south
30550    1172.006345     TabNN  TabNN-num   south
1        1422.205328     TabPFN TabPFN-num  {north west}
    };
\addplot[dashed, thick, const plot mark left, no markers, color=black] coordinates {
    (1, 1422.205328)
    (49692, 1519.050647)
};
\addlegendentry{Pareto frontier}
\end{axis}
\end{tikzpicture}
    \vspace{-0.5em}
    \caption{Models on the \textit{preprocessed} representation.}
    \label{sfig:params_performance_tab}
\end{subfigure}
\hfill
\begin{subfigure}[b]{0.45\textwidth}
    \centering
    \input{figs/model_colors}
\begin{tikzpicture}[scale=0.95]
\begin{axis}[%
axis lines=left,
xmin=0.0,
xmax=1e9,
ymax=1500,
ymin=800,
xmajorgrids,
ymajorgrids,
legend style={at={(0.05,0.85)}, anchor=west, font=\scriptsize},
xlabel={Number of tuned parameters},
ylabel={Mean Elo score},
ylabel near ticks,
label style={font=\small}, 
x tick label style={font=\footnotesize},
y tick label style={font=\footnotesize},
xmode=log,
log basis x={10},
scatter/classes={%
    LR={mark=*,draw=cLR,fill=cLR!50},%
    RF={mark=*,draw=cRF,fill=cRF!50},%
    XGB={mark=*,draw=cXGB,fill=cXGB!50},%
    TabPFN={mark=*,draw=cTabPFN,fill=cTabPFN!50},%
    TabNN={mark=*,draw=cTabNN,fill=cTabNN!50},%
    MBERT={mark=triangle*,draw=cMBERT,fill=cMBERT!50},%
    CORN={mark=triangle*,draw=cCORN,fill=cCORN!50},%
    CLS={mark=triangle*,draw=cCLS,fill=cCLS!50},%
    RAG={mark=triangle*,draw=cRAG,fill=cRAG!50},%
    LoRA={mark=triangle*,draw=cLoRA,fill=cLoRA!50}}]
\addplot[scatter,only marks,mark size=2.6pt,
nodes near coords style = {font=\scriptsize, text=black},
point meta=explicit symbolic,
nodes near coords*={\Method},
visualization depends on={value \thisrow{method} \as \Method},
visualization depends on={value \thisrow{anchor} \as \Anchor},
nodes near coords style = {anchor=\Anchor},
scatter src=explicit symbolic,
forget plot]%
table[meta=label] {
x       y               label   method      anchor
1829        1200.091601  LR     LR-txt      {south east}
32481       1000.000000  RF     RF-txt      {south west}
45000       1128.353243  XGB    XGB-txt     south
67404       919.961906   TabNN  TabNN-txt   west
1           1069.456112  TabPFN TabPFN-txt  {north west}
149608517   1308.047636  MBERT  MBERT       south
149607748   1162.102681  CORN   MBERT-CORN  north
12160       1007.307451  CLS    LLM-CLS     {north east}
12160       911.012289   RAG    LLM-RAG     north
33042304    1226.119088  LoRA   LLM-LoRA    east
    };
\addplot[dashed, thick, const plot mark left, no markers, color=black] coordinates {
    (1, 1069.456112)
    (1829, 1200.091601)
    (33042304, 1226.119088)
    (149608517, 1308.047636)
};
\addlegendentry{Pareto frontier}
\end{axis}
\end{tikzpicture}
    \vspace{-0.5em}
    \caption{Models on the \textit{raw} representation.}
    \label{sfig:params_performance_txt}
\end{subfigure}
\vspace{-0.3em}
\caption{Pareto frontier of mean Elo score versus number of tuned parameters for within-state QR prediction. The dashed line marks the Pareto-optimal set. Parameter counts are averaged over the 12 per-state models. For tree ensembles we count leaves. TabPFN updates no weights and is plotted at $10^0$.}
\label{fig:pareto_within_state}
\end{figure}

For the tabular classifiers on the \textit{preprocessed} representation (Figure~\ref{sfig:params_performance_tab}), the Pareto frontier has only two points. TabPFN-num reaches an Elo score of 1422.2 with zero tuned parameters, because it learns in context. It dominates LR-num, RF-num, and TabNN-num, which all score lower while tuning more parameters. XGB-num is the only model that improves on it. It gains 96.8 Elo points (a 63.6\% expected win rate) with about $5.0\times10^4$ tuned parameters. Figure~\ref{sfig:params_performance_txt} shows the frontier for the classifiers on the \textit{raw} representation. TabPFN-txt again starts the frontier at zero tuned parameters, with an Elo score of 1069.5. LR-txt adds about $1.8\times10^3$ tuned parameters and gains 130.6 points (68.0\%). LLM-LoRA tunes about four orders of magnitude more parameters than LR-txt but adds only 26.0 points (53.7\%). MBERT, the last point on the frontier, finetunes all of its parameters. It tunes about 4.5 times as many parameters as LLM-LoRA and adds another 81.9 points (61.6\%). Lastly, MBERT-CORN tunes the same backbone with one fewer output in its head. However, it scores 64.0 points below LLM-LoRA and is not on the frontier.

\subsubsection{Discussion}
For within-state QR prediction, \textbf{tabular classifiers on the \textit{preprocessed} representation are the recommended choice for the best predictive performance}. They hold the four highest Elo scores, the best QWK in 9 of 12 states, and the best BA in all 12. On average, they also gain 303.5 Elo points over the same model on the \textit{raw} representation. These results are consistent with prior work showing that tree-based models remain state of the art on medium-sized tabular data~\cite{grinsztajn2022tree}. Prior work also shows that text models that split numbers into sub-word tokens represent numerical values less precisely~\cite{wallace2019numbers}. The Pareto analysis adds an efficiency perspective to this recommendation. TabPFN-num ranks second overall with no tuned parameters and outperforms both LR-num and RF-num, so it is a strong default when training resources are scarce. XGB-num leads both the Elo ranking and the mean QWK, so it is the better choice when accuracy matters most. When only the \textit{raw} text is available, MBERT is the strongest option. It beats LLM-LoRA in 17 of the 24 state-metric comparisons, even though its backbone is about 27 times smaller. On the other hand, its ordinal variant, MBERT-CORN, is less reliable within a state. It has the best QWK in MD and WI, but its QWK falls close to zero in MT (0.78), the smallest state, and in GA (4.69), the only three-level state. LR-txt is a strong low-cost alternative, with an Elo score close to that of LLM-LoRA at a small fraction of the tuned parameters. Importantly, these recommendations describe a representative set of model families under the single, uniform configuration of our benchmark. Because no method is tuned per state (Section~\ref{sec:limitations}), they rank the families as evaluated here rather than bound what any of them can reach.

Whether the evaluated models can already \textbf{serve as reliable surrogate QR predictors} depends on how well they agree with human raters. In a few states, the strongest models reach what is commonly interpreted as substantial agreement (61--80)~\cite{landis1977measurement}. In OK, TabPFN-num attains a QWK of 77.34 and XGB-num 76.44. In MD, MBERT-CORN and MBERT attain 70.77 and 70.63. XGB-num's 61.73 in NC sits at the lower edge of this band. At these levels, model predictions could already support decision-making, for example as a monitoring tool. However, in WA, MT, NE, and GA, even the best QWK stays below 35 (33.63, 31.67, 29.99, and 23.33, respectively). This is only fair agreement, with limited practical use for high-stakes decisions. Current models are therefore not yet ready to generate reliable surrogate ratings in every state.

\subsection{Driving Factor Analysis}
\label{sec:results-shap}
QRIS are built not only to rate child care providers but also to help them improve. Predictive accuracy alone does not tell providers what to act on. To connect ratings with improvement, state agencies and providers need to know which features push a provider's rating up or down. Feature attribution can turn black-box predictions into this kind of actionable guidance. It shows each state which factors drive its quality ratings. States can then check whether those factors are the ones linked to children's outcomes.

\subsubsection{Analysis Setup}
We conduct the driving factor analysis with SHapley Additive exPlanations (SHAP)~\cite{lundberg2017unified,lundberg2020}. A SHAP value measures how much a feature contributes to one provider's prediction, relative to the average prediction in that state. We compute exact SHAP values, average their absolute values across providers, and rank features by this mean. The ranking therefore describes what the model relies on across a state's entire provider population.

We explain XGBoost, since XGB-num performs consistently well across the within-state experiments (Section~\ref{sec:results-within}). We make two deliberate changes from the benchmarked configuration. First, the explained model (XGB-reg) is fit with a regression objective that treats the rating as a number, rather than with a multi-class classification objective. It otherwise uses the same \textit{preprocessed} representation and the same configuration as XGB-num (Table~\ref{tab:hparams}). A regression model yields one SHAP value per feature that directly reflects its effect on the predicted rating, whereas a multi-class classifier would produce a separate SHAP value for each rating level. Second, the explained model is fit on all rated providers in a state rather than within five-fold cross-validation, so that it represents the full dataset. Because the explained model is not the one whose accuracy is reported in the within-state benchmark, we also run it through the same five-fold protocol and compare its QWK with that of the benchmarked classifier.

To separate real feature contributions from noise, we implement Boruta's shadow features~\cite{kursa2010}. Each shadow feature is a shuffled copy of a real column. It keeps the column's marginal distribution but breaks any association with the rating. The noise floor is the largest mean absolute SHAP value among all shadow features, and we interpret only the features whose attribution exceeds it. Because tree models tend to favor high-cardinality continuous features over binary ones~\cite{strobl2007bias}, we sample shadow features by cardinality rather than uniformly at random. For each state, we report how many features clear the noise floor out of all features in its \textit{preprocessed} representation.

\subsubsection{Analysis Results}
Table~\ref{tab:shap_combined} reports the fidelity check, which compares the five-fold QWK of XGB-reg with that of XGB-num, the classifier used in the within-state benchmark (Table~\ref{tab:within}). The differences range from $-5.83$ (MT) to $+3.86$ (NE) percentage points. XGB-reg scores higher in 9 of the 12 states and by 0.83 points on average. Only MT shows an absolute difference above 5 points.

\begin{table}[!htbp]
\centering
\footnotesize
\setlength{\tabcolsep}{4pt}
\caption{Fidelity of the XGBoost regression model (XGB-reg) used for SHAP analysis. $\Delta$ is its QWK difference from the XGBoost classifier (XGB-num). Above floor counts the features that exceed the shadow-feature noise floor, and Total counts the features of the \textit{preprocessed} representation.}
\label{tab:shap_combined}
\begin{tabular}{@{}lccccc@{}}
\toprule
\textbf{State} & \multicolumn{3}{c}{\textbf{QWK (\%)}} & \multicolumn{2}{c}{\textbf{\#Features}} \\
\cmidrule(lr){2-4} \cmidrule(lr){5-6}
 & XGB-num& XGB-reg& $\Delta$& Above floor & Total \\
\midrule
CA & 42.59 & 45.85 & $+$3.25 & 4 & 82 \\
CO & 57.98 & 58.95 & $+$0.97 & 7 & 117 \\
GA & 19.00 & 18.26 & $-$0.74 & 5 & 278 \\
KY & 48.10 & 49.65 & $+$1.55 & 9 & 112 \\
MD & 65.23 & 67.95 & $+$2.72 & 3 & 34 \\
MT & 26.63 & 20.80 & $-$5.83 & 0 & 20 \\
NC & 61.73 & 63.28 & $+$1.55 & 5 & 30 \\
NE & 29.99 & 33.86 & $+$3.86 & 5 & 87 \\
OK & 76.44 & 77.34 & $+$0.90 & 7 & 26 \\
SC & 42.18 & 45.53 & $+$3.36 & 2 & 54 \\
WA & 33.63 & 30.58 & $-$3.06 & 5 & 107 \\
WI & 52.63 & 54.01 & $+$1.38 & 6 & 36 \\
\bottomrule
\end{tabular}

\end{table}

\begin{figure}[!htbp]
\centering
\includegraphics[width=\textwidth]{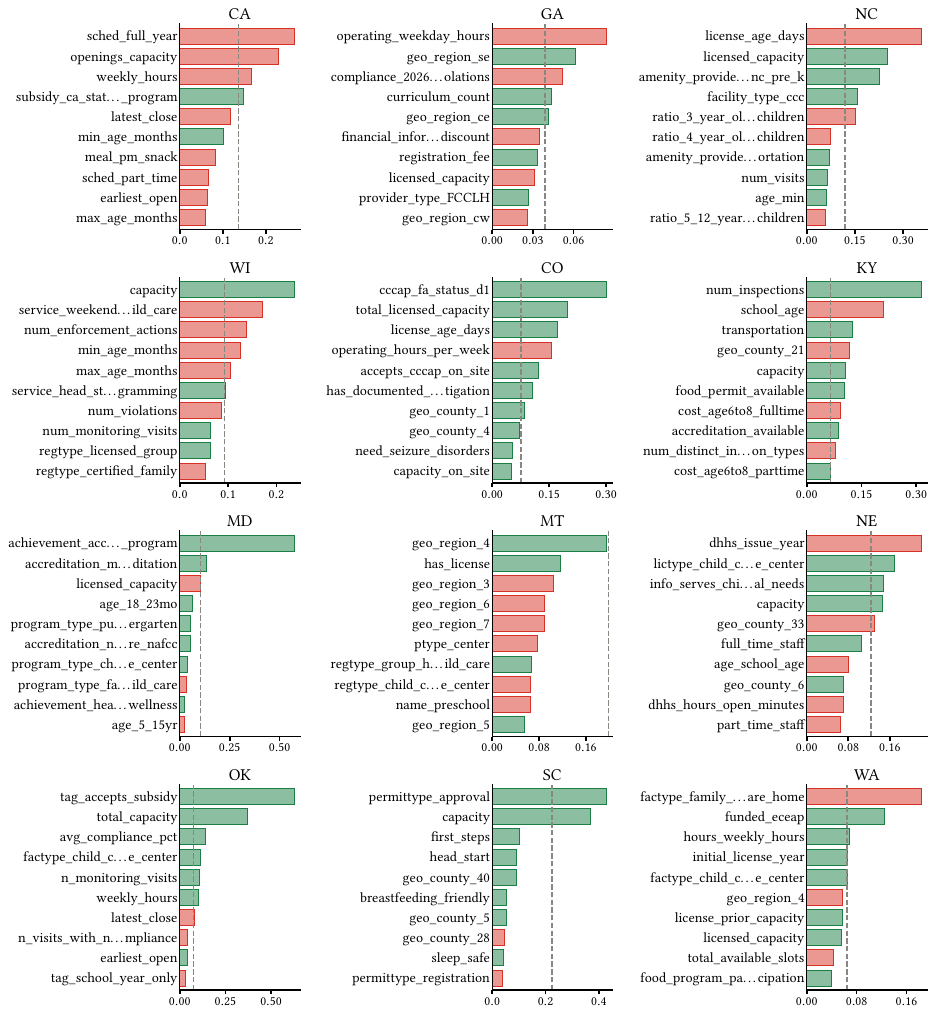}
\caption{Ten highest-attribution features per state, ranked by mean $|\text{SHAP}|$ over that state's providers. The dashed line is the shadow-feature noise floor, and bars to its left are indistinguishable from shuffled noise. Color gives the direction: green where higher feature values push the predicted rating up, red where they push it down. Feature names are state-specific and horizontal scales differ across panels, so bar lengths are comparable within a panel but not across panels.}
\label{fig:shap_grid}
\end{figure}

Figure~\ref{fig:shap_grid} shows the ten highest-attribution features in each state. Across all states, 70 of these 120 features (58.3\%) are positive contributors, whose higher values push the predicted rating up. The other 50 (41.7\%) are negative contributors. Several states are dominated by positive contributors. CO, SC, MD, OK, and WA have 9, 8, 7, 7, and 7 positive features in their top ten. On the other hand, CA, MT, and WI lean negative, with only 2, 4, and 4 positive features.

Table~\ref{tab:shap_combined} also reports how many features in each state clear the shadow-feature noise floor, out of the total available. The count ranges from 0 (MT) to 9 (KY), with 58 of 983 features overall. The dashed line in each panel of Figure~\ref{fig:shap_grid} marks the noise floor. We take the direction of a feature from the correlation between its value and its attribution, which is also what the panel colors encode. Of the 58 features, 37 push the predicted rating up and 21 push it down. Three of them have a correlation below 0.03 in absolute value and are better read as undirected. These are the cost of full-time care at ages 6--8 in KY, licensed capacity in MD, and weekly hours in WA. The 58 features fall into six groups: capacity, operating schedule, and cost (18); program type, licensure, and accreditation (10); geographic and license-administration identifiers (9); services and population served (9); compliance and monitoring (7); and subsidy or public-program participation (5).

\textbf{Institutional formalization is the clearest positive pattern}. Center-based program type clears the floor and is positive in four states, and licensed capacity is positive in seven. All five subsidy or public-funding indicators that clear the floor are positive, as are the public pre-K indicator in NC and Head Start programming in WI. The clearest negative signals are rule violations in GA, enforcement actions in WI, weekend care in WI, and family child care homes in WA. The last is the only home-based indicator that clears the floor. A higher compliance rate in OK raises the predicted rating, which fits this pattern. However, monitoring records are mixed. More inspections in KY, more monitoring visits in OK, and a documented injury investigation in CO all raise the predicted rating, while more distinct inspection types in KY lower it. \textbf{Signals also do not generalize across states.} Operating hours lower the predicted rating in CA, CO, and GA but raise it in OK. License age is positive in CO but negative in NC. Lastly, MT is the only state in which no feature clears the noise floor. Its top feature, a regional indicator, falls just short of it (0.195 against 0.198). When the MT analysis is repeated with 30 different random seeds, this indicator clears the floor in 18 runs, so we treat MT's attribution as inconclusive rather than absent.

\subsubsection{Discussion}
The fidelity check supports reading the SHAP rankings as explanations of the benchmarked classifier. XGB-reg differs from XGB-num by less than 5 QWK points in 11 of 12 states, and it is slightly more accurate on average. MT is the exception, with a gap of $-5.83$ points and an unstable noise floor. MT has only 181 rated providers and 20 binary features in its \textit{preprocessed} representation. The balanced class weights give each of its six one-star providers about 12 times the weight of a two-star provider. A shuffled column can therefore fit a few heavily weighted providers about as well as a real feature can. Without class weights, the regional indicator clears the floor in 29 of 30 seeded runs. We therefore read MT's attributions as inconclusive and draw no conclusions from them.

In the other states, a common pattern emerges. Larger, center-based, publicly funded, and accredited providers tend to receive higher predicted ratings, while rule violations, enforcement actions, and some less formal care arrangements lower them. This is a tendency rather than a rule, as it is state specific. Overall, this pattern \textbf{aligns with child care regulations and existing studies}. \textit{Compliance} indicators provide the clearest alignment. Prior work finds that higher-quality programs generally have fewer licensing violations, although the pattern does not hold in every study or state~\cite{lin2024violations}. Compliance with key health and safety licensing indicators has also been linked to positive child outcomes~\cite{fiene2002indicators}. The positive monitoring records in KY, OK, and CO do not necessarily contradict this. They may reflect exposure to oversight rather than lower quality, for example if larger programs are inspected more often. \textit{Program type and accreditation} features also align with well-documented quality differences. Accreditation is a common marker of quality. For example, NAEYC accreditation is widely relied on as a standard of quality and indicates better-than-average care~\cite{whitebook2004improving}. Similarly, subsidy acceptance is a positive contributor in several states. This is consistent with federal rules that require subsidized providers to meet baseline health and safety standards, and with the higher subsidy rates that many states pay for higher quality~\cite{ccdf2016rule}.

\textit{Capacity and operational scale} present a more nuanced picture. Larger licensed capacity is generally a positive contributor. This is in line with evidence of economies of scale in center-based care, where centers serving more children operate at a lower cost per child without an apparent loss of quality~\cite{helburn1996child}. However, weekend care and long operating hours are negative contributors in some states. This contrasts with the finding that centers with longer hours of operation had lower costs per child with no apparent ill effects on quality~\cite{helburn1996child}. Lastly, geographic identifiers also surface above the noise floor. This suggests that unmeasured regional factors, such as local economic conditions, child care supply, or differences in how regulations are implemented, may influence ratings in ways that our dataset only partially captures.

\subsection{Cross-State QR Prediction Experiment}
\label{sec:results-cross}
Table~\ref{tab:loso_full} reports the leave-one-state-out transfer results at every target supervision level. Figure~\ref{fig:elo_cross_state} summarizes overall cross-state performance with Elo scores computed from BA and QWK across all targets and supervision levels. All models consume the \textit{raw} representation, so every tabular classifier appears in its \textit{-txt} form.

As described in Section~\ref{sssec:models}, the one-shot-fit models are fit once on the union of the pooled source states and the $p\%$ target draw. The sequentially trained models are first pretrained on the pooled source states and then finetuned on the target draw. Each phase derives its own class weights and early-stopping split. At $p{=}0$, there is no target data to finetune on, so the zero-shot result of a sequentially trained model is its source-only pretraining result. Lastly, each experiment uses the target state's native rating scale. Georgia is therefore left out of the five-level source pools. When Georgia is the target, the five-level source states are collapsed onto the reconciled three-level label space as $\{1,2\}{\to}1$, $\{3\}{\to}2$, and $\{4,5\}{\to}3$.

\begin{table}[!htbp]
\centering
\scriptsize
\setlength{\tabcolsep}{3pt}
\caption{Leave-one-state-out cross-state QR prediction results. At every target supervision level $p$, each cell is scored on the same stratified 20\% test set of the held-out target state. The best value per state and metric within each supervision level, excluding Dummy, is in \textbf{bold}, and the second best is \underline{underlined}. Dummy predicts the majority class of its training set (the pooled source states plus the target draw).}
\label{tab:loso_full}
\resizebox{\textwidth}{!}{%
\begin{tabular}{l|cc|cc|cc|cc|cc|cc|cc|cc|cc|cc|cc|cc}
\toprule
\multirow{2}{*}{\textbf{Method}} & \multicolumn{2}{c|}{\textbf{CA}} & \multicolumn{2}{c|}{\textbf{CO}} & \multicolumn{2}{c|}{\textbf{GA}} & \multicolumn{2}{c|}{\textbf{KY}} & \multicolumn{2}{c|}{\textbf{MD}} & \multicolumn{2}{c|}{\textbf{MT}} & \multicolumn{2}{c|}{\textbf{NC}} & \multicolumn{2}{c|}{\textbf{NE}} & \multicolumn{2}{c|}{\textbf{OK}} & \multicolumn{2}{c|}{\textbf{SC}} & \multicolumn{2}{c|}{\textbf{WA}} & \multicolumn{2}{c}{\textbf{WI}}\\
 & BA & QWK & BA & QWK & BA & QWK & BA & QWK & BA & QWK & BA & QWK & BA & QWK & BA & QWK & BA & QWK & BA & QWK & BA & QWK & BA & QWK\\
\midrule
\multicolumn{25}{c}{\textit{0\% target supervision}}\\
\midrule
Dummy & 20.00 & 0.00 & 20.00 & 0.00 & 33.33 & 0.00 & 20.00 & 0.00 & 20.00 & 0.00 & 20.00 & 0.00 & 20.00 & 0.00 & 20.00 & 0.00 & 20.00 & 0.00 & 20.00 & 0.00 & 0.00 & 0.00 & 0.00 & 0.00\\
LR-txt & 14.77 & -3.88 & 22.38 & \textbf{17.31} & 32.00 & 2.06 & 20.48 & 5.80 & 23.04 & 35.10 & 24.71 & 5.19 & 17.78 & -5.89 & 18.35 & 11.36 & 20.51 & 1.06 & \textbf{26.79} & \textbf{21.66} & \textbf{27.12} & 1.01 & 25.92 & 7.60\\
RF-txt & 13.53 & -9.36 & 17.58 & 2.62 & 32.92 & -0.55 & 14.51 & 2.41 & 27.73 & \underline{48.96} & 19.00 & -8.22 & 23.26 & -0.57 & 19.66 & 4.57 & 16.61 & -9.08 & \underline{24.99} & \underline{20.88} & 22.41 & 3.87 & 23.34 & \textbf{12.52}\\
XGB-txt & \textbf{41.28} & \underline{-1.22} & 18.67 & -0.03 & 29.30 & -6.32 & 17.74 & 2.75 & 23.35 & 40.56 & \textbf{29.36} & -1.44 & \textbf{36.48} & \underline{23.06} & 18.21 & 3.28 & 14.57 & -7.02 & 20.12 & -0.70 & 21.50 & 4.65 & \textbf{27.86} & 3.43\\
TabPFN-txt & 13.85 & -11.21 & \textbf{23.20} & 5.76 & \underline{38.09} & 6.20 & 21.53 & -0.72 & \textbf{33.66} & 33.03 & 27.00 & \textbf{9.09} & \underline{31.60} & \textbf{28.63} & 20.15 & 4.12 & 20.30 & -0.46 & 20.00 & 0.00 & 22.74 & -5.43 & 24.91 & 2.15\\
TabNN-txt & 9.91 & -7.80 & 22.15 & 2.03 & 33.31 & 1.42 & 19.66 & -3.15 & 27.99 & \textbf{49.08} & \underline{28.29} & 7.59 & 18.77 & 0.19 & 20.81 & -5.18 & \textbf{26.48} & \textbf{27.69} & 20.00 & 0.00 & 22.90 & -3.67 & 14.21 & 6.65\\
MBERT & 18.18 & -1.61 & 18.72 & 10.29 & \textbf{39.67} & \textbf{11.61} & \underline{21.87} & 1.49 & 25.19 & 46.06 & 14.29 & 0.00 & 17.60 & 5.95 & \underline{24.44} & 10.67 & \underline{21.33} & \underline{15.49} & 20.00 & 2.50 & 17.16 & 1.00 & 25.37 & 0.63\\
MBERT-CORN & 17.14 & -8.00 & 20.00 & 1.11 & 34.94 & 2.70 & 18.24 & 0.54 & 25.18 & 20.10 & 18.43 & 2.78 & 11.70 & -0.42 & 24.25 & \underline{14.26} & 19.87 & -2.14 & 23.30 & 7.64 & \underline{25.26} & \underline{10.75} & \underline{27.25} & 5.17\\
LLM-CLS & \underline{20.88} & -6.14 & 13.81 & -6.70 & 35.28 & 6.32 & 20.00 & -0.19 & 25.00 & 45.63 & 22.50 & -4.42 & 28.60 & 3.95 & 15.94 & 3.00 & 18.48 & -0.99 & 20.29 & -12.23 & 23.96 & -2.19 & 11.67 & -6.61\\
LLM-RAG & 18.46 & -7.24 & \underline{22.60} & \underline{11.10} & 35.33 & \underline{10.58} & 21.12 & \underline{8.74} & 21.97 & 46.18 & 23.93 & 6.86 & 17.22 & 1.07 & 20.00 & 5.24 & 15.75 & 0.66 & 20.00 & 0.00 & 23.03 & 1.54 & 25.12 & \underline{7.79}\\
LLM-LoRA & 19.58 & \textbf{-0.28} & 18.33 & 5.85 & 32.46 & -0.13 & \textbf{27.39} & \textbf{9.25} & \underline{29.67} & 32.79 & 11.43 & \underline{9.02} & 25.05 & 13.14 & \textbf{32.39} & \textbf{24.13} & 20.43 & 14.38 & 20.60 & 0.10 & 21.63 & \textbf{14.04} & 21.99 & -2.73\\
\midrule
\multicolumn{25}{c}{\textit{16\% target supervision}}\\
\midrule
Dummy & 20.00 & 0.00 & 20.00 & 0.00 & 33.33 & 0.00 & 20.00 & 0.00 & 20.00 & 0.00 & 20.00 & 0.00 & 20.00 & 0.00 & 20.00 & 0.00 & 20.00 & 0.00 & 20.00 & 0.00 & 0.00 & 0.00 & 0.00 & 0.00\\
LR-txt & 24.44 & 9.10 & 27.90 & 31.28 & 35.21 & 1.63 & 21.33 & 6.86 & 41.45 & 68.77 & 21.93 & -2.41 & 25.04 & 19.42 & 22.44 & 12.84 & 34.88 & 55.42 & 24.85 & 11.13 & 27.36 & 5.51 & 38.42 & 40.08\\
RF-txt & 24.54 & 1.00 & 28.65 & 34.89 & 34.02 & -3.60 & 20.54 & 0.69 & 41.79 & 67.84 & \underline{41.57} & \underline{30.14} & 31.49 & 39.28 & 29.49 & 12.51 & 35.67 & 58.10 & \textbf{37.39} & \textbf{35.95} & 25.12 & -2.08 & 31.95 & 25.07\\
XGB-txt & 22.44 & 3.12 & 30.41 & 36.54 & 34.15 & 2.62 & 21.86 & 3.91 & 41.52 & 67.23 & \textbf{45.00} & \textbf{37.84} & 29.13 & 34.01 & 28.79 & 8.76 & 38.49 & 59.08 & 33.27 & \underline{33.84} & 26.47 & 6.07 & 33.82 & 27.90\\
TabPFN-txt & 20.03 & 2.11 & 29.89 & 42.19 & 34.14 & 0.13 & 20.00 & 0.00 & 41.64 & 68.58 & 28.57 & 5.63 & 25.28 & 18.05 & 22.97 & 9.46 & 33.31 & 59.45 & 20.26 & -4.20 & 25.00 & 0.00 & 30.35 & 21.86\\
TabNN-txt & 19.24 & 2.47 & 23.66 & 20.47 & 28.09 & -7.88 & 22.88 & 13.73 & 36.90 & 62.64 & 26.29 & 10.69 & 25.16 & 25.50 & 15.50 & 1.25 & 26.39 & 18.69 & 20.00 & 0.00 & 33.80 & 19.04 & 29.80 & 14.52\\
MBERT & \underline{31.54} & \underline{30.83} & \textbf{32.50} & \underline{48.38} & \textbf{40.86} & \textbf{15.78} & \underline{32.02} & \underline{23.41} & \textbf{44.72} & \underline{71.68} & 33.29 & 10.94 & \underline{36.20} & \underline{47.03} & 34.23 & \textbf{36.58} & \underline{42.55} & 58.19 & \underline{37.27} & 29.17 & \underline{36.32} & \underline{30.15} & \underline{39.41} & \underline{42.07}\\
MBERT-CORN & 26.74 & 19.20 & 29.77 & \textbf{50.50} & \underline{38.43} & \underline{14.49} & 22.36 & 19.12 & \underline{44.25} & \textbf{72.11} & 15.36 & 13.79 & \textbf{40.23} & \textbf{52.03} & \underline{34.59} & 29.29 & \textbf{45.20} & \textbf{70.33} & 19.45 & 31.89 & \textbf{36.82} & 29.95 & \textbf{41.10} & \textbf{51.70}\\
LLM-CLS & 24.61 & 2.94 & 27.61 & 30.94 & 34.06 & 0.62 & 22.40 & 3.50 & 43.52 & 68.72 & 17.00 & 0.00 & 33.85 & 32.59 & 30.85 & 21.78 & 27.96 & 24.50 & 28.75 & 7.33 & 33.53 & 19.33 & 33.86 & 27.58\\
LLM-RAG & 26.79 & 21.91 & \underline{31.00} & 31.38 & 35.40 & 3.68 & 22.66 & 16.51 & 42.60 & 68.16 & 14.00 & 5.41 & 31.83 & 43.66 & 24.55 & 2.73 & 20.44 & 1.71 & 28.39 & 15.04 & 26.41 & 5.54 & 30.65 & 23.75\\
LLM-LoRA & \textbf{33.23} & \textbf{36.44} & 29.58 & 37.60 & 36.80 & 10.59 & \textbf{38.55} & \textbf{35.21} & 44.10 & 70.93 & 39.29 & 28.07 & 35.53 & 41.56 & \textbf{37.96} & \underline{31.87} & 42.30 & \underline{63.07} & 37.02 & 31.66 & 35.49 & \textbf{34.99} & 35.14 & 39.10\\
\midrule
\multicolumn{25}{c}{\textit{32\% target supervision}}\\
\midrule
Dummy & 20.00 & 0.00 & 20.00 & 0.00 & 33.33 & 0.00 & 20.00 & 0.00 & 20.00 & 0.00 & 20.00 & 0.00 & 20.00 & 0.00 & 20.00 & 0.00 & 20.00 & 0.00 & 20.00 & 0.00 & 0.00 & 0.00 & 0.00 & 0.00\\
LR-txt & 22.81 & 5.15 & 29.76 & 38.89 & 34.66 & 0.62 & 21.32 & 7.52 & 41.58 & 68.95 & \underline{28.71} & 10.96 & 26.00 & 28.81 & 25.06 & 14.75 & 37.86 & 60.01 & 26.94 & 18.96 & 28.28 & 10.21 & 38.14 & 37.09\\
RF-txt & 26.42 & 20.00 & 30.37 & 36.74 & 36.03 & 5.94 & 21.96 & 9.56 & 42.79 & 67.97 & 28.57 & 24.32 & 32.13 & 43.54 & 27.41 & 9.05 & 41.18 & 61.66 & \underline{37.60} & 42.58 & 26.88 & 1.05 & 33.61 & 27.70\\
XGB-txt & 27.52 & 20.28 & \underline{32.34} & 43.35 & 35.60 & 4.50 & 24.67 & 18.74 & 42.67 & 69.51 & \textbf{30.50} & \underline{31.58} & 31.62 & 41.50 & 28.80 & 9.63 & 41.25 & 63.19 & \textbf{41.97} & \textbf{50.80} & 29.88 & 12.97 & 35.08 & 36.45\\
TabPFN-txt & 20.00 & 0.00 & 29.74 & 42.22 & 34.10 & 0.05 & 19.91 & 0.60 & 40.03 & 68.70 & 18.57 & -5.06 & 25.76 & 24.53 & 24.30 & -2.11 & 35.89 & 61.98 & 30.50 & \underline{44.43} & 25.00 & 0.00 & 31.22 & 24.37\\
TabNN-txt & 24.20 & 18.39 & 28.31 & 24.94 & 31.02 & -0.11 & 24.38 & 14.78 & 38.42 & 66.46 & 15.14 & -2.63 & 32.88 & 16.55 & 24.32 & -1.75 & 40.40 & 57.82 & 19.68 & 1.09 & 25.83 & 4.07 & 32.06 & 24.24\\
MBERT & 27.61 & \underline{32.53} & 32.28 & \underline{44.06} & \textbf{44.40} & 16.46 & \textbf{39.53} & \textbf{38.09} & \underline{43.92} & \underline{71.70} & 19.00 & 16.44 & \textbf{38.50} & \textbf{56.48} & \underline{31.86} & 15.64 & \underline{52.45} & \underline{74.39} & 36.38 & 37.87 & \underline{36.07} & \underline{28.45} & \underline{41.94} & \underline{47.85}\\
MBERT-CORN & \underline{35.62} & \textbf{41.39} & 31.67 & \textbf{48.61} & 41.73 & \textbf{21.43} & 27.13 & 30.73 & 41.25 & 71.34 & 26.71 & \textbf{40.74} & 27.76 & \underline{45.10} & 27.90 & \underline{19.73} & 50.54 & \textbf{76.88} & 29.44 & 39.25 & 34.02 & 24.26 & \textbf{44.10} & \textbf{50.71}\\
LLM-CLS & \textbf{47.90} & 24.17 & 27.10 & 31.26 & 38.58 & 12.48 & 27.66 & 20.27 & \textbf{44.09} & 70.92 & 18.21 & -17.91 & 29.72 & 29.99 & 24.55 & 5.56 & 30.27 & 35.93 & 33.05 & 28.37 & \textbf{39.26} & \textbf{31.12} & 35.43 & 31.00\\
LLM-RAG & 20.98 & 1.39 & 29.16 & 39.16 & 40.39 & 10.22 & 27.14 & 26.96 & 41.70 & 71.23 & 18.00 & 6.06 & \underline{34.72} & 44.46 & 27.29 & 7.54 & 28.25 & 38.02 & 25.16 & 11.44 & 35.56 & 28.12 & 30.52 & 22.40\\
LLM-LoRA & 27.49 & 27.50 & \textbf{34.49} & 43.68 & \underline{42.99} & \underline{16.86} & \underline{38.34} & \underline{36.62} & 43.07 & \textbf{72.11} & 18.93 & 14.63 & 32.33 & 39.63 & \textbf{33.08} & \textbf{26.47} & \textbf{54.61} & 72.66 & 34.02 & 28.46 & 35.47 & 27.54 & 40.23 & 44.89\\
\midrule
\multicolumn{25}{c}{\textit{48\% target supervision}}\\
\midrule
Dummy & 20.00 & 0.00 & 20.00 & 0.00 & 33.33 & 0.00 & 20.00 & 0.00 & 20.00 & 0.00 & 20.00 & 0.00 & 20.00 & 0.00 & 20.00 & 0.00 & 20.00 & 0.00 & 20.00 & 0.00 & 0.00 & 0.00 & 0.00 & 0.00\\
LR-txt & 22.63 & 8.83 & 31.19 & 40.55 & 35.53 & 3.24 & 22.55 & 10.63 & 41.26 & 68.75 & 27.79 & 15.19 & 26.02 & 30.82 & 26.28 & 17.32 & 38.02 & 60.44 & 28.64 & 25.09 & 28.95 & 11.62 & 38.10 & 38.20\\
RF-txt & 26.45 & 15.06 & 29.21 & 36.37 & 36.52 & 6.86 & 26.86 & 24.51 & 42.75 & 68.41 & 29.93 & 29.27 & 32.94 & 39.28 & 30.52 & 14.20 & 39.09 & 61.54 & 40.36 & 51.46 & 28.38 & 4.79 & 34.95 & 32.50\\
XGB-txt & 30.06 & 29.90 & 30.73 & 44.18 & 35.69 & 5.30 & 26.26 & 23.81 & 42.40 & 69.02 & \underline{30.43} & \textbf{36.59} & 30.71 & 38.66 & 29.38 & \underline{20.21} & 39.79 & 65.16 & 43.49 & \textbf{57.15} & 30.14 & 10.92 & 36.54 & 36.00\\
TabPFN-txt & 21.06 & 4.42 & 31.00 & 42.55 & 34.38 & 0.78 & 21.24 & 15.07 & 40.11 & 68.99 & 20.00 & 10.13 & 26.59 & 29.82 & 26.78 & 19.00 & 33.95 & 61.20 & 33.50 & 40.80 & 26.81 & 3.69 & 33.49 & 32.45\\
TabNN-txt & 20.03 & 11.09 & 31.71 & 32.33 & 35.67 & 1.84 & 24.86 & 21.06 & 38.25 & 65.53 & 12.50 & 0.00 & 38.17 & 43.30 & 33.60 & 1.51 & 39.03 & 61.59 & 22.85 & 3.22 & 37.05 & 16.87 & 36.49 & 31.50\\
MBERT & 35.68 & \underline{39.93} & \underline{33.59} & 47.87 & \textbf{40.92} & \underline{16.80} & 36.20 & \underline{34.66} & \underline{44.66} & \underline{71.30} & 24.29 & 33.90 & \textbf{39.98} & \textbf{55.89} & \textbf{36.63} & \textbf{28.50} & \underline{54.26} & 72.33 & \textbf{45.13} & 51.59 & 40.27 & \textbf{34.88} & \textbf{45.32} & \underline{52.49}\\
MBERT-CORN & \underline{37.97} & \textbf{47.58} & 31.87 & \textbf{52.99} & \underline{40.50} & \textbf{16.97} & \textbf{38.81} & 33.64 & 40.26 & 71.12 & 23.79 & 19.67 & \underline{38.63} & \underline{53.56} & 29.72 & 18.91 & 50.85 & \textbf{75.55} & \underline{43.87} & \underline{54.18} & 40.10 & \underline{34.35} & \underline{43.17} & \textbf{57.70}\\
LLM-CLS & \textbf{48.02} & 28.04 & 31.84 & 37.33 & 39.95 & 11.21 & 32.61 & 27.44 & 41.26 & 70.94 & 22.64 & 0.00 & 33.97 & 45.90 & 27.31 & 17.43 & 34.66 & 38.99 & 28.10 & 6.49 & \textbf{43.05} & 28.27 & 37.96 & 36.03\\
LLM-RAG & 28.15 & 26.53 & 32.16 & 37.94 & 36.58 & 5.42 & 34.42 & 26.37 & 40.33 & 70.35 & 20.00 & 15.05 & 30.81 & 40.93 & 33.12 & 15.15 & 31.73 & 35.68 & 34.62 & 12.74 & 36.89 & 28.46 & 34.28 & 26.20\\
LLM-LoRA & 27.01 & 26.18 & \textbf{37.30} & \underline{48.30} & 40.07 & 14.71 & \underline{38.73} & \textbf{37.58} & \textbf{44.67} & \textbf{71.49} & \textbf{32.14} & \underline{34.67} & 36.39 & 49.43 & \underline{36.19} & 19.66 & \textbf{57.45} & \underline{74.23} & 38.52 & 35.20 & \underline{42.02} & 31.27 & 40.83 & 47.15\\
\midrule
\multicolumn{25}{c}{\textit{64\% target supervision}}\\
\midrule
Dummy & 20.00 & 0.00 & 20.00 & 0.00 & 33.33 & 0.00 & 20.00 & 0.00 & 20.00 & 0.00 & 20.00 & 0.00 & 20.00 & 0.00 & 20.00 & 0.00 & 20.00 & 0.00 & 20.00 & 0.00 & 0.00 & 0.00 & 0.00 & 0.00\\
LR-txt & 21.43 & 7.13 & 30.69 & 43.51 & 35.13 & 3.28 & 23.47 & 14.99 & 41.44 & 68.97 & \underline{32.79} & 19.75 & 26.45 & 31.58 & 26.12 & 14.26 & 38.60 & 61.43 & 28.64 & 26.46 & 29.78 & 10.29 & 38.27 & 39.74\\
RF-txt & 25.13 & 25.09 & 30.12 & 38.41 & 35.30 & 5.19 & 25.80 & 19.19 & 41.29 & 68.17 & 24.50 & 25.88 & 33.68 & 40.60 & 28.35 & 14.13 & 38.33 & 61.23 & 40.62 & \underline{52.35} & 27.97 & 7.23 & 35.95 & 32.93\\
XGB-txt & 34.02 & 36.52 & 32.44 & 42.72 & 35.56 & 7.24 & 26.94 & 22.57 & 41.72 & 69.24 & 26.50 & 21.69 & 31.65 & 37.94 & 30.77 & 24.57 & 40.40 & 63.71 & \textbf{42.50} & \textbf{54.45} & 29.27 & 10.00 & 36.35 & 36.97\\
TabPFN-txt & 21.75 & 9.33 & 30.77 & 43.28 & 33.88 & 1.92 & 20.65 & 13.99 & 41.62 & 69.20 & 22.14 & 8.79 & 26.65 & 29.32 & 27.80 & 11.16 & 34.34 & 59.96 & 35.51 & 43.40 & 27.17 & 3.37 & 33.69 & 33.25\\
TabNN-txt & 20.25 & 13.41 & 30.01 & 31.88 & 37.09 & 2.96 & 26.52 & 18.13 & \underline{45.26} & 69.66 & 14.93 & 13.91 & 36.52 & 48.56 & 24.57 & 12.80 & 39.24 & 56.28 & 29.24 & 32.03 & 35.34 & 20.61 & 34.64 & 26.13\\
MBERT & 34.96 & 39.33 & 33.08 & \underline{46.80} & \textbf{43.62} & \underline{18.85} & \textbf{42.47} & \textbf{39.59} & \textbf{45.68} & 71.69 & \textbf{36.21} & \textbf{42.42} & \underline{38.23} & \underline{50.10} & \textbf{39.68} & \underline{27.20} & \textbf{57.12} & \underline{77.74} & 41.38 & 46.80 & \underline{40.52} & 32.40 & 40.36 & \underline{48.13}\\
MBERT-CORN & 37.16 & \textbf{46.25} & \underline{34.31} & \textbf{54.71} & \underline{40.69} & \textbf{21.96} & \underline{41.81} & \underline{35.58} & 40.72 & \textbf{71.92} & 18.21 & \underline{36.07} & 37.90 & \textbf{57.81} & 33.15 & 18.57 & 51.06 & \textbf{78.29} & 35.98 & 50.89 & 39.90 & \underline{37.04} & \underline{44.02} & \textbf{55.04}\\
LLM-CLS & \textbf{52.76} & 35.44 & 26.71 & 34.28 & 34.57 & 5.06 & 24.17 & 20.97 & 42.80 & \underline{71.75} & 27.36 & 0.00 & 34.23 & 48.06 & \underline{35.27} & 18.88 & 34.07 & 38.26 & 27.92 & 2.87 & 38.43 & 28.40 & 40.56 & 38.90\\
LLM-RAG & 29.87 & 16.98 & 26.83 & 26.16 & 36.58 & 4.88 & 28.22 & 26.84 & 39.53 & 70.20 & 22.50 & -5.06 & 32.34 & 44.50 & 27.09 & 13.87 & 27.68 & 28.71 & 32.37 & 27.61 & \textbf{40.96} & 21.87 & 37.49 & 26.89\\
LLM-LoRA & \underline{39.70} & \underline{39.35} & \textbf{36.50} & 45.61 & 40.26 & 17.14 & 40.32 & 32.73 & 43.98 & 71.26 & 30.07 & 23.38 & \textbf{38.41} & 49.69 & 34.97 & \textbf{37.87} & \underline{55.02} & 76.70 & \underline{42.34} & 51.86 & 38.16 & \textbf{43.97} & \textbf{44.57} & 46.23\\
\midrule
\multicolumn{25}{c}{\textit{80\% target supervision}}\\
\midrule
Dummy & 20.00 & 0.00 & 20.00 & 0.00 & 33.33 & 0.00 & 20.00 & 0.00 & 20.00 & 0.00 & 20.00 & 0.00 & 20.00 & 0.00 & 20.00 & 0.00 & 20.00 & 0.00 & 20.00 & 0.00 & 0.00 & 0.00 & 0.00 & 0.00\\
LR-txt & 26.49 & 14.58 & 31.51 & 43.57 & 35.20 & 1.50 & 23.73 & 15.05 & 41.62 & 68.66 & 30.79 & 19.51 & 26.20 & 31.50 & 25.16 & 12.05 & 40.65 & 63.10 & 29.98 & 31.08 & 30.45 & 11.45 & 39.28 & 42.55\\
RF-txt & 24.63 & 27.15 & 32.12 & 42.72 & 34.44 & -0.93 & 27.21 & 19.43 & 41.92 & 67.05 & 22.00 & 12.66 & 35.09 & 48.01 & 28.82 & 14.95 & 41.80 & 63.26 & 40.12 & \textbf{49.84} & 28.43 & 6.90 & 36.46 & 34.24\\
XGB-txt & 33.02 & 36.15 & 32.56 & 45.39 & 35.36 & 3.63 & 28.08 & 19.23 & 43.12 & 68.15 & 29.00 & 21.18 & 33.18 & 46.35 & 29.19 & 24.33 & 43.77 & 67.00 & \underline{41.41} & \underline{48.53} & 30.46 & 16.70 & 37.37 & 38.60\\
TabPFN-txt & 24.55 & 18.22 & 30.74 & 43.40 & 33.96 & -0.10 & 22.32 & 10.56 & 41.29 & \underline{69.16} & 23.57 & 7.06 & 26.24 & 32.78 & 26.37 & 6.93 & 34.13 & 61.80 & 27.76 & 24.99 & 27.17 & 3.37 & 34.52 & 34.97\\
TabNN-txt & 29.66 & 10.67 & 29.39 & 36.25 & 35.45 & 4.93 & 32.52 & 23.23 & \textbf{47.58} & 65.52 & \textbf{33.00} & -2.99 & \underline{38.50} & 32.48 & 27.18 & 14.97 & 42.00 & 59.40 & 26.88 & 25.44 & 33.50 & 7.79 & 37.93 & 33.13\\
MBERT & 30.88 & 31.09 & \textbf{37.36} & \textbf{51.06} & \textbf{44.61} & \textbf{20.67} & 36.24 & 36.20 & 44.75 & 68.75 & 26.57 & \underline{32.43} & 37.59 & 51.86 & 35.34 & 24.96 & \textbf{57.85} & \textbf{78.75} & \textbf{43.85} & 47.89 & \underline{43.11} & \underline{42.84} & \underline{46.69} & \underline{57.36}\\
MBERT-CORN & \underline{38.01} & \underline{39.42} & 33.34 & \underline{49.87} & 40.88 & 18.15 & \textbf{46.32} & \textbf{42.64} & 41.10 & \textbf{71.92} & 24.21 & 18.18 & 30.55 & \textbf{54.12} & \underline{40.74} & \underline{29.95} & \underline{56.20} & \underline{78.72} & 37.44 & 47.03 & \textbf{45.46} & \textbf{44.78} & \textbf{47.03} & \textbf{58.59}\\
LLM-CLS & 25.93 & 5.98 & 28.98 & 36.59 & 41.00 & 12.60 & 34.39 & 24.56 & 43.57 & 66.75 & 23.57 & 17.07 & 29.33 & 42.13 & 32.75 & 24.64 & 34.12 & 40.97 & 18.13 & -8.95 & 36.55 & 28.39 & 42.02 & 42.26\\
LLM-RAG & 30.06 & \textbf{40.03} & 30.21 & 39.74 & 41.10 & 16.49 & 29.38 & 25.01 & \underline{47.13} & 68.09 & \underline{32.14} & 8.51 & 35.46 & 49.87 & 31.47 & 14.56 & 30.18 & 39.20 & 40.12 & 32.08 & 35.98 & 29.11 & 39.43 & 34.91\\
LLM-LoRA & \textbf{38.60} & 35.87 & \underline{36.27} & 47.25 & \underline{41.58} & \underline{18.51} & \underline{39.81} & \underline{38.24} & 41.49 & 68.91 & 26.57 & \textbf{32.50} & \textbf{39.28} & \underline{53.46} & \textbf{42.44} & \textbf{39.40} & 55.05 & 74.76 & 39.91 & 34.85 & 40.20 & 36.49 & 45.53 & 49.31\\
\bottomrule
\end{tabular}}

\end{table}

\subsubsection{Predictive Performance}
\label{sssec:cross-overall}
MBERT attains the highest Elo score (1374.4), followed by LLM-LoRA (1324.9) and MBERT-CORN (1298.6) (Figure~\ref{fig:elo_cross_state}). XGB-txt follows at 1106.7. LLM-RAG (1025.8), LLM-CLS (1014.9), RF-txt (1000.0), and LR-txt (977.9) cluster around the anchor, ahead of TabNN-txt (892.1) and TabPFN-txt (876.0). The Dummy baseline scores 594.9. The 267.7-point gap between MBERT and XGB-txt corresponds to an 82.4\% expected win rate. In contrast, the 49.5-point gap between MBERT and LLM-LoRA corresponds to only 57.1\%.

In the zero-shot setting ($p{=}0$), most models perform close to chance in 11 of the 12 targets. Across the 120 non-Dummy cells at this level, the median QWK is 2.66, 41 cells are at or below zero, and only 16 exceed 20. MD accounts for 10 of these 16 cells. Every non-Dummy model exceeds a QWK of 20 on MD, with a median of 43.10 and a maximum of 49.08 (TabNN-txt). In the other eleven targets, the median zero-shot QWK lies between $-6.69$ (CA) and 4.91 (NE). Averaged over the twelve targets, the best zero-shot model is LLM-LoRA with a mean QWK of 9.96, and no model reaches 10.

Once target supervision is available, the finetuned language models take the lead. At $p{=}16$, MBERT, MBERT-CORN, or LLM-LoRA holds the best QWK in 10 of the 12 targets. The exceptions are MT, where XGB-txt leads, and SC, where RF-txt leads. The three models also hold the best QWK in at least 10 of 12 targets at every higher level. Among them, MBERT-CORN wins most often. It has the best QWK in 28 of the 60 combinations of target and supervision level with $p>0$, against 12 for MBERT and 11 for LLM-LoRA. It also has the highest mean QWK at $p{=}80$, with 46.11 against 45.32 for MBERT and 44.13 for LLM-LoRA. The remaining models trail well behind at that level: XGB-txt at 36.27, LLM-RAG at 33.13, RF-txt at 32.11, LR-txt at 29.55, LLM-CLS at 27.75, TabPFN-txt at 26.10, and TabNN-txt at 25.90. However, MBERT-CORN's lead is specific to QWK. On BA, MBERT beats it in 42 of these 60 combinations, which helps explain why MBERT ranks higher by Elo. This split is consistent with CORN's ordinal objective~\cite{shi2023corn}, which targets the ordering of ratings that QWK rewards rather than per-class accuracy. Lastly, the GA target, which tests transfer across rating scales, stays low throughout. Its best QWK at any supervision level is 21.96 (MBERT-CORN, $p{=}64$).

\subsubsection{Effect of Target Supervision}
\label{sssec:cross-supervision}
\begin{figure}[!htbp]
    \centering
    \begin{subfigure}[b]{0.48\textwidth}
        \centering
    \input{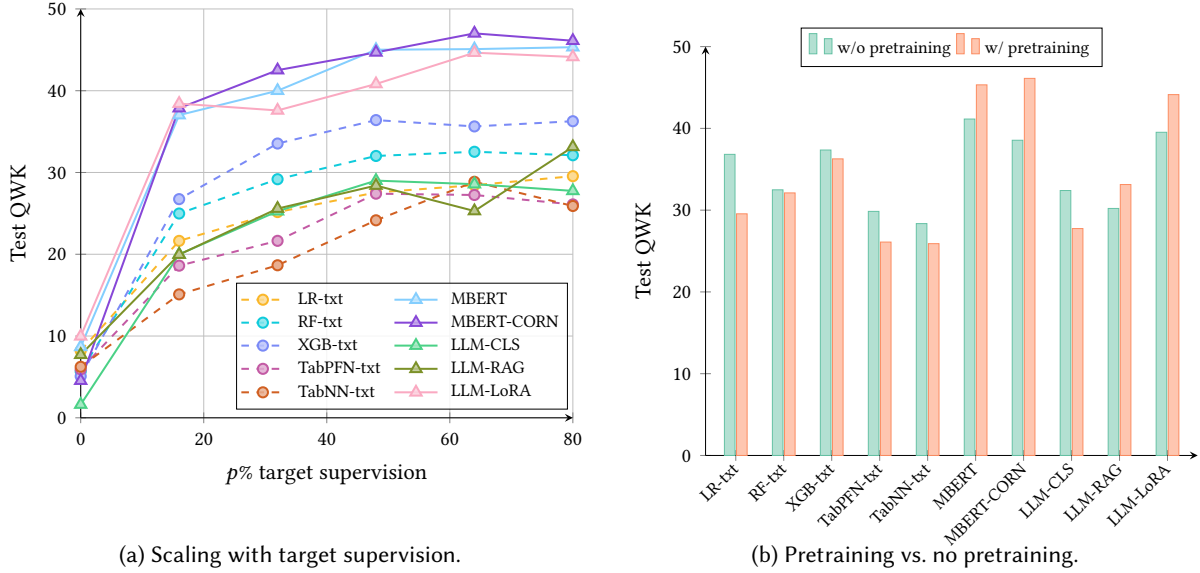}
\begin{tikzpicture}[scale=0.95]
\begin{axis}[%
axis lines=left,
xmin=0,
xmax=80,
ymax=50,
ymin=0,
xmajorgrids,
ymajorgrids,
xlabel={$p$\% target supervision},
ylabel={Test QWK},
x tick label style={font=\footnotesize},
y tick label style={font=\footnotesize},
ylabel near ticks,
label style={font=\small},
legend style={at={(0.99,0.02)}, anchor=south east, font=\scriptsize, legend columns=5,
  row sep=-1pt, column sep=3pt, inner sep=1.5pt, fill=white, fill opacity=0.85, text opacity=1},
legend cell align=left,
transpose legend,
]
\addplot[color=cLR, dashed, mark=*, mark size=2pt, mark options={solid, fill=cLR!50, draw=cLR}, thick] coordinates {
  (0, 8.198)
  (16, 21.636)
  (32, 25.160)
  (48, 27.557)
  (64, 28.449)
  (80, 29.550)
};
\addlegendentry{LR-txt}
\addplot[color=cRF, dashed, mark=*, mark size=2pt, mark options={solid, fill=cRF!50, draw=cRF}, thick] coordinates {
  (0, 5.671)
  (16, 24.983)
  (32, 29.176)
  (48, 32.021)
  (64, 32.533)
  (80, 32.107)
};
\addlegendentry{RF-txt}
\addplot[color=cXGB, dashed, mark=*, mark size=2pt, mark options={solid, fill=cXGB!50, draw=cXGB}, thick] coordinates {
  (0, 5.083)
  (16, 26.743)
  (32, 33.542)
  (48, 36.408)
  (64, 35.635)
  (80, 36.270)
};
\addlegendentry{XGB-txt}
\addplot[color=cTabPFN, dashed, mark=*, mark size=2pt, mark options={solid, fill=cTabPFN!50, draw=cTabPFN}, thick] coordinates {
  (0, 5.930)
  (16, 18.605)
  (32, 21.642)
  (48, 27.408)
  (64, 27.248)
  (80, 26.095)
};
\addlegendentry{TabPFN-txt}
\addplot[color=cTabNN, dashed, mark=*, mark size=2pt, mark options={solid, fill=cTabNN!50, draw=cTabNN}, thick] coordinates {
  (0, 6.238)
  (16, 15.093)
  (32, 18.654)
  (48, 24.153)
  (64, 28.863)
  (80, 25.902)
};
\addlegendentry{TabNN-txt}
\addplot[color=cMBERT, solid, mark=triangle*, mark size=2.6pt, mark options={solid, fill=cMBERT!50, draw=cMBERT}, thick] coordinates {
  (0, 8.673)
  (16, 37.017)
  (32, 39.997)
  (48, 45.012)
  (64, 45.088)
  (80, 45.322)
};
\addlegendentry{MBERT}
\addplot[color=cCORN, solid, mark=triangle*, mark size=2.6pt, mark options={solid, fill=cCORN!50, draw=cCORN}, thick] coordinates {
  (0, 4.541)
  (16, 37.867)
  (32, 42.514)
  (48, 44.685)
  (64, 47.011)
  (80, 46.114)
};
\addlegendentry{MBERT-CORN}
\addplot[color=cCLS, solid, mark=triangle*, mark size=2.6pt, mark options={solid, fill=cCLS!50, draw=cCLS}, thick] coordinates {
  (0, 1.619)
  (16, 19.986)
  (32, 25.263)
  (48, 29.006)
  (64, 28.572)
  (80, 27.749)
};
\addlegendentry{LLM-CLS}
\addplot[color=cRAG, solid, mark=triangle*, mark size=2.6pt, mark options={solid, fill=cRAG!50, draw=cRAG}, thick] coordinates {
  (0, 7.710)
  (16, 19.957)
  (32, 25.583)
  (48, 28.402)
  (64, 25.287)
  (80, 33.133)
};
\addlegendentry{LLM-RAG}
\addplot[color=cLoRA, solid, mark=triangle*, mark size=2.6pt, mark options={solid, fill=cLoRA!50, draw=cLoRA}, thick] coordinates {
  (0, 9.963)
  (16, 38.424)
  (32, 37.587)
  (48, 40.822)
  (64, 44.649)
  (80, 44.129)
};
\addlegendentry{LLM-LoRA}
\end{axis}
\end{tikzpicture}
    \caption{Scaling with target supervision.}
    \label{fig:target_supervision_scaling}
    \end{subfigure}
    \hfill
    \begin{subfigure}[b]{0.48\textwidth}
    \begin{tikzpicture}[scale=0.95]
\begin{axis}[
    ybar=0.6pt,
    cycle list/Set2,
    axis lines=left,
    every axis plot post/.style={fill=.!50},
    symbolic x coords={LR-txt,RF-txt,XGB-txt,TabPFN-txt,TabNN-txt,MBERT,MBERT-CORN,LLM-CLS,LLM-RAG,LLM-LoRA},
    xtick=data,
    x tick label style={yshift=-2pt, rotate=45, anchor=east, inner sep=0pt, font=\scriptsize},
    ylabel={Test QWK},
    ylabel near ticks,
    label style={font=\small}, 
    y tick label style={font=\footnotesize},
    ymin=0, ymax=50,
    bar width=4.2pt,
    enlarge x limits=0.07,
    legend style={at={(0.5,0.95)}, anchor=south, legend columns=-1, font=\scriptsize},
]

\addplot+[ybar] coordinates {(LR-txt, 36.822) (RF-txt, 32.485) (XGB-txt, 37.353) (TabPFN-txt, 29.848) (TabNN-txt, 28.354) (MBERT, 41.141) (MBERT-CORN, 38.547) (LLM-CLS, 32.402) (LLM-RAG, 30.208) (LLM-LoRA, 39.521)};
\addplot+[ybar] coordinates {(LR-txt, 29.550) (RF-txt, 32.107) (XGB-txt, 36.270) (TabPFN-txt, 26.095) (TabNN-txt, 25.902) (MBERT, 45.322) (MBERT-CORN, 46.114) (LLM-CLS, 27.749) (LLM-RAG, 33.133) (LLM-LoRA, 44.129)};
\legend{w/o pretraining, w/ pretraining}
\end{axis}
\end{tikzpicture}
    \vspace{-2em}
    \caption{Pretraining vs.\ no pretraining.}
    \label{fig:rq3_pretrain_compare}
    \end{subfigure}
    \caption{(a) Mean test QWK over the twelve target states as a function of the target supervision level $p$ for all ten models. (b) Mean test QWK of the same ten models over the twelve states, with and without pretraining on the pooled source states. ``w/ pretraining'' is the $p{=}80$ cross-state setting. ``w/o pretraining'' is the within-state setting.}
\end{figure}

Figure~\ref{fig:target_supervision_scaling} plots the mean QWK over the twelve targets against the supervision level $p$ for all ten models. For every model, the largest gain occurs between $p{=}0$ and $p{=}16$. The three finetuned language models rise the most in this first step: MBERT-CORN from 4.54 to 37.87, LLM-LoRA from 9.96 to 38.42, and MBERT from 8.67 to 37.02. XGB-txt rises from 5.08 to 26.74, RF-txt from 5.67 to 24.98, LR-txt from 8.20 to 21.64, LLM-CLS from 1.62 to 19.99, LLM-RAG from 7.71 to 19.96, TabPFN-txt from 5.93 to 18.61, and TabNN-txt from 6.24 to 15.09. Relative to the total gain from $p{=}0$ to $p{=}80$, this first step accounts for 83\% of the gain for LLM-LoRA, 80\% for MBERT-CORN, 77\% for MBERT, 73\% for RF-txt, 70\% for LLM-CLS, 69\% for XGB-txt, and 63\% for both LR-txt and TabPFN-txt. LLM-RAG (48\%) and TabNN-txt (45\%) are the exceptions and gain more gradually.

Beyond $p{=}16$, the other eight curves flatten. Their additional gain from $p{=}16$ to $p{=}80$ ranges from 5.71 for LLM-LoRA to 9.53 for XGB-txt. Only MBERT and LR-txt rise at every step. LLM-LoRA dips at $p{=}32$ and ends below its $p{=}64$ value. XGB-txt, TabPFN-txt, and LLM-CLS peak at $p{=}48$, and MBERT-CORN and RF-txt peak at $p{=}64$. In contrast, LLM-RAG's curve is the least regular. It falls from 28.40 at $p{=}48$ to 25.29 at $p{=}64$ and then rises to 33.13 at $p{=}80$. TabNN-txt does the reverse, falling from 28.86 at $p{=}64$ to 25.90 at $p{=}80$.

\subsubsection{Discussion}
The cross-state results lead to three conclusions. First, \textbf{zero-shot transfer did not prove viable for cross-state QR prediction} in our benchmark, with one exception. Without target supervision, the best source-trained model exceeds a QWK of 20 in only five of the twelve targets: MD (49.08), NC (28.63), OK (27.69), NE (24.13), and SC (21.66). In the other seven, it stays below 18. MD is the only target on which transfer works across models. It is also the state whose within-state attribution is dominated by accreditation flags that are tied to the rating by policy (Section~\ref{sec:results-shap}). One plausible reading is that this policy link, rather than a general quality signal, is what transfers to MD. On this evidence, a state without any rated providers should not expect a usable predictor from other states' data alone. 

Second, \textbf{the amount of target supervision needed is modest}. Labeling 16\% of a target state's providers delivers 63\% to 83\% of the gain from full supervision for eight of the ten models, and the returns diminish after that. At this level, the mean QWK of MBERT-CORN, LLM-LoRA, and MBERT already reaches 98\%, 97\%, and 90\% of their within-state means, although the two settings are scored on different test sets (Section~\ref{sec:results-pretrain}). A state that rates a stratified sample of its providers can therefore bootstrap a predictor for the rest at a fraction of the assessment cost.

Third, \textbf{the recommended models reverse relative to the within-state setting}. Since the state-specific \textit{preprocessed} schemas cannot be pooled, cross-state prediction is limited to the \textit{raw} representation. On this representation, the finetuned language models (MBERT, MBERT-CORN, and LLM-LoRA) lead the Elo ranking and hold the best QWK in at least 10 of 12 targets at every supervision level above zero. Within this group, MBERT-CORN is the strongest choice when ordinal agreement matters most, and MBERT is the most balanced across both metrics. Tree ensembles on sentence embeddings remain competitive in a few targets. They hold the best QWK in SC at every supervised level and in MT at two of five, but they are not the general choice. Lastly, the GA target shows the limit of the reconciled label space. Collapsing five levels onto three keeps GA's QWK below 22 at every supervision level. A coarse label mapping alone was therefore not sufficient to bridge rating systems with different scales, and transfer across rating scales remains an open problem.

\subsection{Transfer Utility Analysis}
\label{sec:results-pretrain}
Cross-state QR prediction measures what a target state can achieve by borrowing data from other states. A complementary question concerns a state that already has a substantial labeled set. Does pretraining on other states add anything beyond training on its own data alone? To answer it, we compare the within- and cross-state settings. For each of the twelve states, the model without pretraining is the within-state model, which sees only the target state's own providers. The model with pretraining is the same model pretrained on the pooled source states and then adapted on the target state's full training pool ($p{=}80$). Both settings therefore train on 80\% of the target state's rated providers. We include both one-shot-fit and sequentially trained models, so the comparison also shows how each training regime uses source data. Georgia is included on its native three-level scale in both settings, with the pooled source states coarsened to that scale. For each model, we report the mean test QWK across the twelve states in the two settings.

\subsubsection{Analysis Results}
Figure~\ref{fig:rq3_pretrain_compare} reports the mean test QWK of all ten models over the twelve states, with pretraining (``w/ pretraining'') and without it (``w/o pretraining''). The effect of pretraining depends on the training regime. Four of the six sequentially trained models gain from pretraining. MBERT-CORN gains the most, rising from 38.55 to 46.11 ($+7.56$). It is followed by LLM-LoRA from 39.52 to 44.13 ($+4.61$), MBERT from 41.14 to 45.32 ($+4.18$), and LLM-RAG from 30.21 to 33.13 ($+2.92$). The gain is most consistent for MBERT-CORN, LLM-LoRA, and MBERT, which improve in 11, 10, and 9 of the 12 states, while LLM-RAG improves in only 6. Part of MBERT-CORN's gain repairs its weakest within-state results, with rises of 17.4 points in MT and 13.5 in GA. However, sequential training alone does not guarantee a gain. LLM-CLS falls from 32.40 to 27.75 ($-4.65$), and TabNN-txt from 28.35 to 25.90 ($-2.45$). For the four one-shot-fit models, which are fit on the union of the pooled source states and the target data, adding source data lowers the mean QWK. LR-txt falls from 36.82 to 29.55 ($-7.27$), TabPFN-txt from 29.85 to 26.10 ($-3.75$), XGB-txt from 37.35 to 36.27 ($-1.08$), and RF-txt from 32.49 to 32.11 ($-0.38$). Overall, MBERT-CORN, MBERT, and LLM-LoRA hold the three highest mean QWK values after pretraining.

\subsubsection{Discussion}
The pretraining comparison shows that whether cross-state data help depends less on their availability than on how they are used. The clearest benefit appears for the three language models whose backbone is finetuned: MBERT, MBERT-CORN, and LLM-LoRA. For them, pretraining on the pooled source states before adapting on the target raises the mean QWK by more than four points, even when the target's full training pool is available. In this setup, \textbf{cross-state data add value beyond a state's own labels}, and the gain holds in most states. For the frozen-backbone variants, the result is mixed. LLM-RAG gains 2.92 points, but LLM-CLS, which differs from it only in lacking the exemplars, loses 4.65. One hypothesis is that LLM-RAG's head learns to compare an input with the per-class exemplars. This comparison may transfer across states more readily than the fixed class boundaries that LLM-CLS learns on the source. TabNN-txt, the only sequentially trained tabular classifier, also loses 2.45 points. In contrast, for the one-shot-fit models, \textbf{mixing the pooled source into a single training set lowers performance}. Without a mechanism to prioritize the target distribution, the larger and differently distributed source data appear to dominate the fit. For states that want to draw on other states' ratings, we therefore recommend the sequential pretrain-then-adapt framework with a finetuned language model such as MBERT, MBERT-CORN, or LLM-LoRA.

\section{Conclusion}
We presented \textit{CCQ}, a de-identified dataset of 59,372 child care providers across 12 U.S. states. \textit{CCQ} comes in two releases: a \textit{preprocessed} release for tabular models and a \textit{raw} release for language models. We also release the LLM-based curation pipeline that produced both. Using \textit{CCQ}, we benchmarked tabular and textual classifiers on within- and cross-state QR prediction. In within-state QR prediction, tabular classifiers on the \textit{preprocessed} representation perform best. XGB-num ranks first, and TabPFN-num ranks second without any tuned parameters. However, agreement with true ratings is substantial in only a few states and fair in others, so predictions should support rather than replace on-site assessment. In cross-state QR prediction, zero-shot transfer is close to chance in every target except MD. On the other hand, labeling 16\% of a target state's providers delivers most of the gain from full target supervision, and finetuned language models then perform best. Pretraining on other states also improves these models, even when the target state's full training pool is available. Lastly, feature attribution associates higher predicted ratings with larger, center-based, publicly funded, and accredited providers, and lower ones with rule violations and enforcement actions. We hope \textit{CCQ} will support future work on transfer across rating scales, state-specific adaptation, and the drivers of child care quality.

\subsection{Limitations}
\label{sec:limitations}
Several limitations of the dataset and benchmark should be kept in mind. On the dataset side, our preprocessing is one of several reasonable ways to structure the data. The curation agents were also evaluated only on Georgia, the reference state (Section~\ref{sssec:data_curation-clean}), so their measured accuracy may not carry over to the other eleven states. Since the cleaning scripts for these states were not verified line by line, residual cleaning errors can exist. On the benchmark side, every method runs under a single, uniform configuration, with no per-state tuning and a single random seed. The reported numbers are therefore reference baselines rather than the best attainable performance of any model. Cross-state results are single point estimates on a fixed 20\% test set of the target state. For the smallest state, MT, with 181 rated providers in total, this test set holds only a few dozen providers. Cross-state results for the GA target also depend on one fixed mapping from five rating levels to three. Furthermore, the feature attributions describe what the explained model relies on, not causal effects on quality. Some high-attribution features, such as accreditation in MD and subsidy acceptance in OK, are themselves tied to state policy. Part of their attribution may therefore reflect administrative requirements linked to QR scores rather than an independent quality signal. Lastly, geographic identifiers likely act as proxies for regional factors that the released features do not capture.
\subsection{Ethical Considerations}
\label{sec:ethics}

This study was approved by the Emory University Institutional Review Board (Protocol \#: 2025P012045). The dataset contains only aggregated, facility-level information for each child care provider and no individual-level data. To further protect human subjects, we remove or anonymize every feature that could directly identify a specific provider. This minimizes the risk of re-identification and confidentiality breaches. All data handling and analysis followed the approved IRB protocol and relevant ethical guidelines.

We ask users of \textit{CCQ} to follow three guidelines. First, users should not attempt to re-identify providers, for example by linking released records back to the state portals. Second, predicted ratings are research estimates, not official ratings, and should not inform licensing, funding, or enforcement decisions about individual providers. Third, the records reflect the state portals at the time of collection, and ratings may have changed since.

\begin{acks}
This research was partially supported by the 2025-2026 RISE Pilot Grant Program from the \textit{Center for Children's Health Assessment, Research Translation, and Combating Environmental Risk} (US National Institute of Environmental Health Sciences 1P2CES033430-01).
\end{acks}

\bibliographystyle{ACM-Reference-Format}
\bibliography{refs}

\end{document}